\documentclass[10pt,twocolumn,letterpaper]{article}

 \usepackage{iccv}              
\usepackage{multirow}
\usepackage{utfsym}
\usepackage{colortbl}
\usepackage[titletoc]{appendix}

\usepackage[accsupp]{axessibility}

\definecolor{iccvblue}{rgb}{0.21,0.49,0.74}
\usepackage[pagebackref,breaklinks,colorlinks,allcolors=iccvblue]{hyperref}

\def\paperID{6531} 
\def\confName{ICCV}
\def\confYear{2025}

\title{UniConvNet: Expanding Effective Receptive Field while Maintaining Asymptotically Gaussian Distribution for ConvNets of Any Scale}

\author{Yuhao Wang\\
Xi'an Jiaotong University\\
{\tt\small yuhaowang.ai@gmail.com}
\and
Wei Xi \thanks{Corresponding author}\\
Xi'an Jiaotong University\\
{\tt\small xiwei@xjtu.edu.cn}
}

\begin{document}
\maketitle
\begin{abstract}
	Convolutional neural networks (ConvNets) with large effective receptive field (ERF), still in their early stages, have demonstrated promising effectiveness while constrained by high parameters and FLOPs costs and disrupted asymptotically Gaussian distribution (AGD) of ERF.
	This paper proposes an alternative paradigm: rather than merely employing extremely large ERF, it is more effective and effcient to expand the ERF while maintaining AGD of ERF by proper combination of smaller kernels, such as $7\times{7}$, $9\times{9}$, $11\times{11}$.
	This paper introduces a Three-layer Receptive Field Aggregator and designs a Layer Operator as the fundamental operator from the perspective of receptive field. The ERF can be expanded to the level of existing large-kernel ConvNets through the stack of proposed modules while maintaining AGD of ERF.
	Using these designs, we propose a universal model for ConvNet of any scale, termed UniConvNet.
	Extensive experiments on ImageNet-1K, COCO2017, and ADE20K demonstrate that UniConvNet outperforms state-of-the-art CNNs and ViTs across various vision recognition tasks for both lightweight and large-scale models with comparable throughput. 
	Surprisingly, UniConvNet-T achieves $84.2\%$ ImageNet top-1 accuracy with $30M$ parameters and $5.1G$ FLOPs.
	UniConvNet-XL also shows competitive scalability to big data and large models, acquiring $88.4\%$ top-1 accuracy on ImageNet. Code and models are publicly available at \href{https://github.com/ai-paperwithcode/UniConvNet}{https://github.com/ai-paperwithcode/UniConvNet}.
\end{abstract}

\section{Introduction}
\label{sec:intro}

With the impressive triumph of transformers \cite{ViT, Swin-transformer}, constructing long-range dependencies has become a crucial principle in designing convolutional neural networks (ConvNets). Some prior works \cite{RepLKNet, SLaK, UniRepLKNet} have made attempts to capture relationships across large receptive fields, surpassing traditional convolutional neural networks \cite{ResNet, MobileViTv1, EfficientNet, Shufflenet, RegNet} and achieving significant improvements in various vision recognition tasks, such as image classification, object detection, instance segmentation, and semantic segmentation.
Current ConvNets achieve long-range dependencies by scaling up the convolutional kernel with re-parameterization \cite{RepLKNet,DBB}, parameter sharing \cite{pelk} or sparsity \cite{SLaK} techniques. 
Some recent works leverage the key properties of large kernels \cite{UniRepLKNet} or encode their interactions \cite{moganet} to inform ConvNet architecture design.
They benefit from large ERF while constrained by high parameters and FLOPs costs.

\begin{figure}[t]
	\centering
	\includegraphics[width=\linewidth]{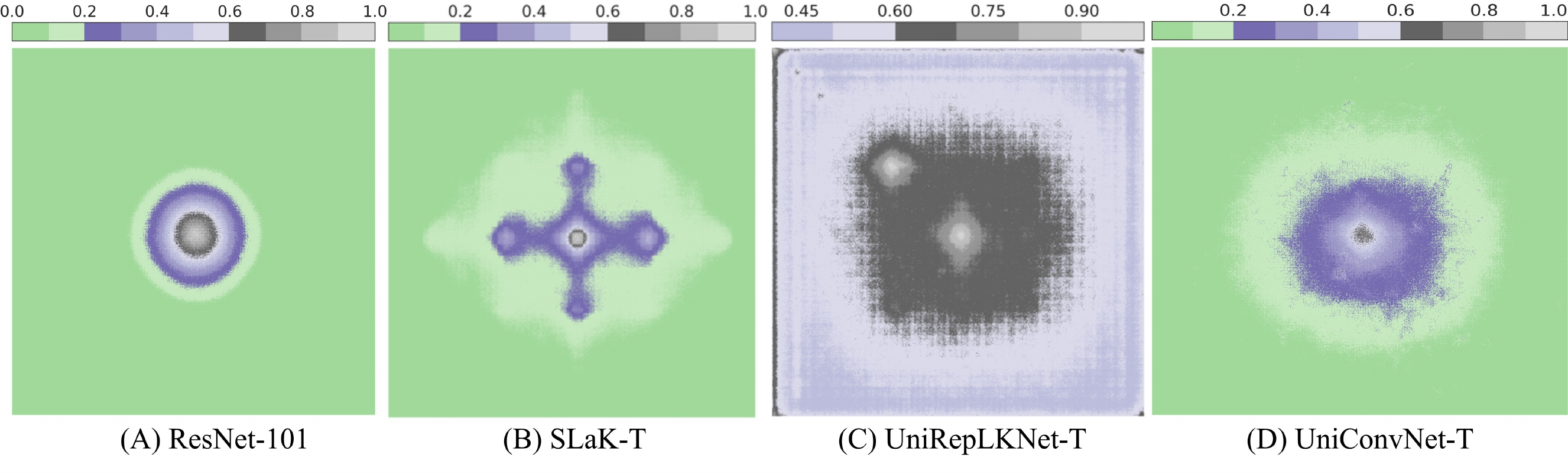}
	
		
		\caption{Effective Receptive Field (ERF) of ResNet-101, SLaK-T, UniRepLKNet-T, UniConvNet-T. The more stepped colour area from the center indicates better asymptotically Gaussian distribution (AGD) of ERF. The wider area indicates a larger ERF. Large-kernel ConvNets, such as SLaK-T and UniRepLKNet-T, disrupt the AGD of ERF.}
	\label{fig:intro}
\end{figure}

A typical paradigm \cite{ResNet, Res2net, ResNeXt} for ConvNets is to use a stack of many small spatial convolutions (\eg 3×3) to enlarge the receptive fields in ConvNets.
Why small-kernel ConvNets constrained by small ERF still get effective performance?
The conventional ConvNets, such as ResNet-101 \cite{ResNet}, have a small ERF but benefit from multi-scale impact (gradient), which follows AGD, through the stack of $3\times{3}$ convolution modules, as shown in \cref{fig:intro} (A). 
This suggests that smaller-scale pixels, around the position of the output pixel, of the input should have more impact on the output pixel.
Large-kernel ConvNets, such as SLaK-T \cite{SLaK} and UniRepLKNet-T \cite{UniRepLKNet}, achieve a large ERF but disrupt the AGD, which either obtain a discriminative impact at weird position or get similar impacts of different scales, as depicted in \cref{fig:intro} (B) and \cref{fig:intro} (C).

\begin{figure*}[t]
	\centering
	\includegraphics[width=\linewidth]{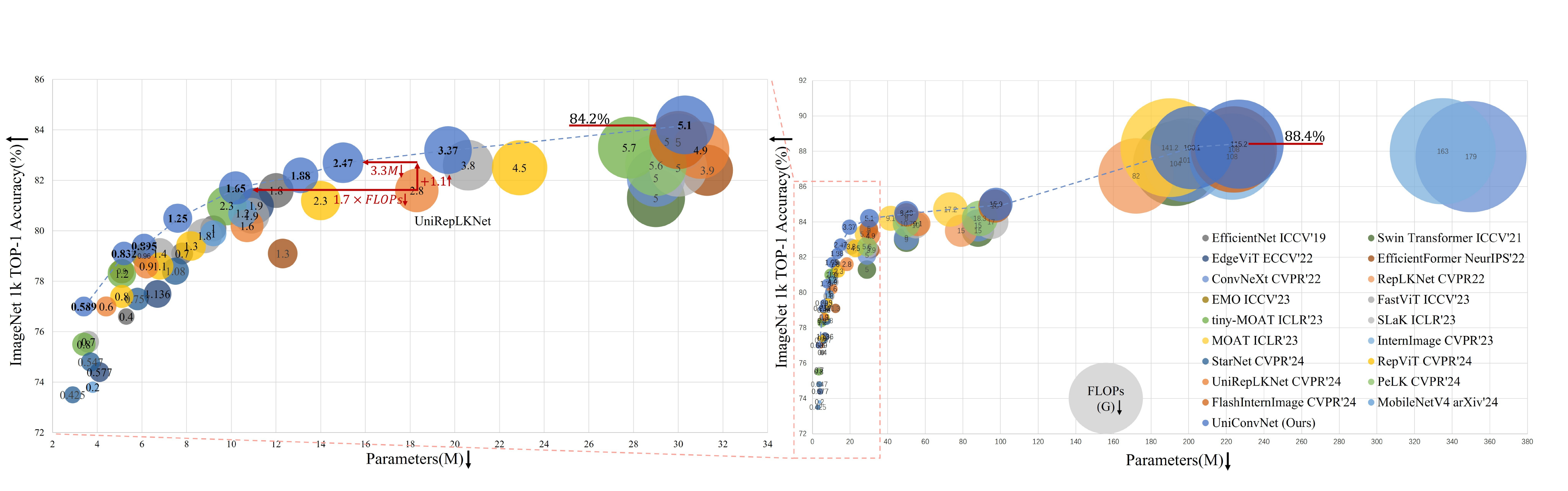}
	
		
		\caption{\textbf{\bfseries{Comparison of parameters and accuracy between UniConvNet (ours) and others.}} The area of the circle, the number in the circle, represents the FLOPs of the relevant model. UniConvNet achieves the best accuracy-parameter and accuracy-FLOPs trade-off.}
	\label{fig:TLDR}
\end{figure*}

\textit{Is there a proper way to combine smaller kernels to expand ERF while maintaining the AGD of ERF?}
This paper proposes an alternative paradigm: rather than merely employing extremely large ERF, it is more effective and efficient to expand the ERF while maintaining the AGD of ERF by proper combination of smaller kernels.

To answer this question, we introduce a Receptive Field Aggregator (RFA) for ConvNets, designed to obtain AGD at a shadow module by directly assign impact for different scales.
The input images are separated into multiple heads according to the layer of RFA.
Parameters and FLOPs costs are reduced by recursively feeding multi-head inputs into layer operators, creating a pyramidal increment among channels. 
For heads with different patterns among channels, in each layer, we propose a spatial encoder from the perspective of receptive field, called the Layer Operator (LO). 
The LO consists of two components: the Amplifier (Amp) and the Discriminator (Dis). 
The Amp expands the scale of the receptive field and amplifies the impacts of pixels on the receptive field by element-wise multiplication. 
The salient pixels in the receptive field will have a more distinguished impact.
The Dis provides small-scale impacts from new pixels to the large receptive field produced by the Amp. 
The final receptive field becomes a large two-layer receptive field.
Sequentially, each LO expands and amplifies the receptive field of the previous LO by Amp and provides a discriminative receptive field for adding impacts of small-scale pixels. 
The final receptive field of three-layer RFA results in a four-layer receptive field following an AGD.
The ERF can be expanded and the AGD of ERF can be maintained through the stack of many RFA modules, as illustrated in \cref{fig:intro} (D).

With these designs, the proposed UniConvNet efficiently reduces parameters and FLOPs while getting a multi-scale impact on ERF.
Consequently, it outperforms state-of-the-art CNNs and ViTs in various vision recognition tasks, from lightweight to large-scale models, as illustrated in \cref{fig:TLDR}.
Notably, UniConvNet-T achieves $84.2\%$ TOP-1 accuracy, surpassing models with similar parameters and FLOPs by at least $0.6$ points, representing a significant improvement over existing ConvNets \cite{ConvNeXt, SLaK, pelk, DCNV4}. UniConvNet-XL breaks through ConvNet bottleneck, achieving $88.4\%$ TOP-1 accuracy with a superior parameters and FLOPs trade-off compared to contemporary CNNs \cite{ConvNeXt,DCNV3,DCNV4,hornet,moganet,RepLKNet} and ViTs \cite{CoAtNet,Swin-transformer,Swin-transformer-v2,MOAT}.
UniConvNet is also powerful on downstream tasks.
UniConvNet-L obtains $55.7\%$ on COCO \cite{COCO} and $55.1\%$ on ADE20K \cite{ADE20K}.

We believe the high performance of UniConvNet is mainly because of the large ERF\cite{ERF} while maintaining AGD as compared in \cref{fig:intro}. 
The ERF scale is comparable with ConvNets using extremly large kernels.
The AGD of ERF are more similar to the intuition that the closer pixels should have more impacts.
We hope our findings can help to understand the intrinsic mechanism of ConvNets.

\section{Model Architecture}
\label{sec:model_arch}


\begin{figure}[t]
	\centering
	\includegraphics[width=\linewidth]{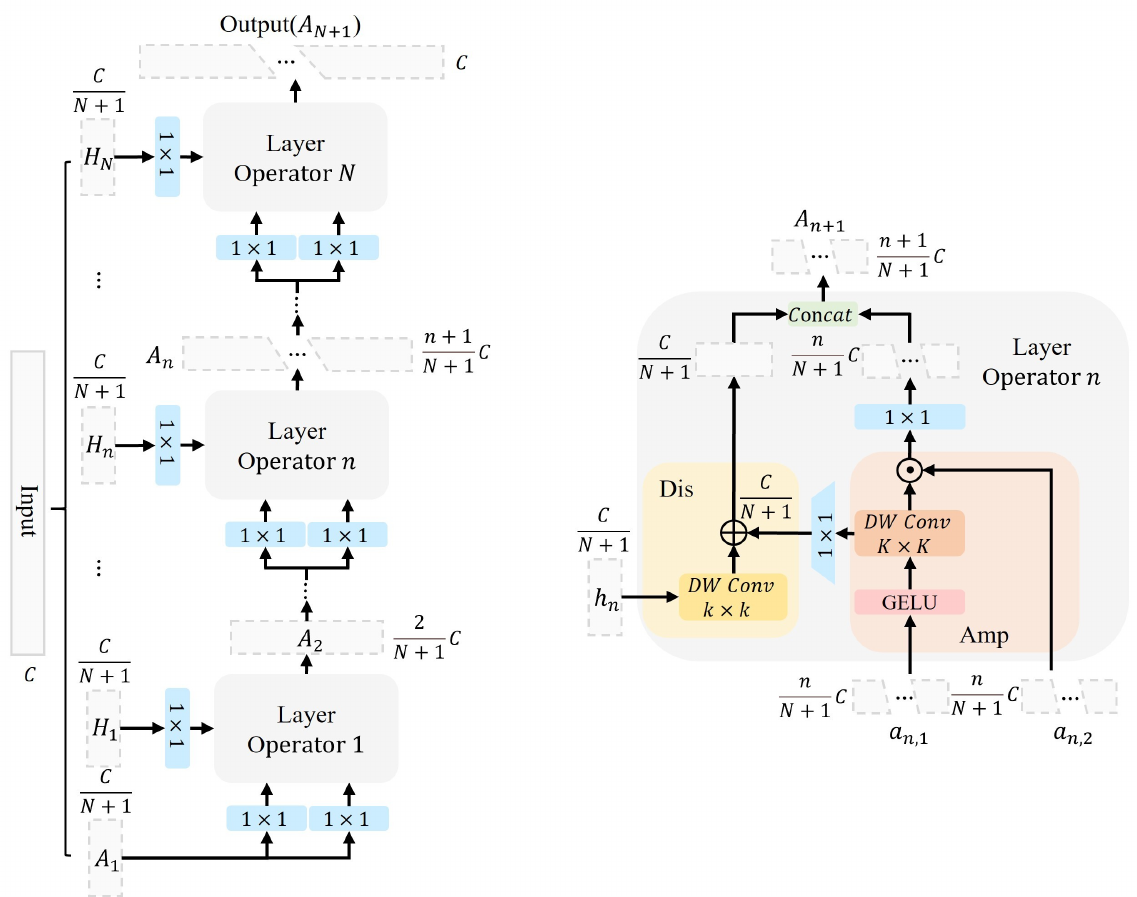}
	
	\caption{\textbf{\bfseries{Left:}} Receptive Field Aggregator. \textbf{\bfseries{Right:}} Layer Operator.}
	\label{fig:Recursive-Convolution}
\end{figure}

\subsection{Rceptive Field Aggregator}

In order to expand the effective receptive field (ERF) \cite{ERF} while maintianning the asymptotically Gaussian distribution (AGD) of ERF, we introduce a Rceptive Field Aggregator (RFA) as illustrated in \cref{fig:Recursive-Convolution} (left), which directly assign discriminative impact on receptive field of different scales at a shadow module.
Specifically, input images are initially divided into $N+1$ parts along the channel dimension based on the layer $N$ of RFA 
resulting in $N+1$ heads: $A_1$, $H_1$, ..., $H_n$, ..., $H_N$.
The input heads can be categorized into two parts: $A_1$ and $H_1$, ..., $H_n$, ..., $H_N$. 
Concretely, the input head $A_1 \in \mathbb{R}^{B \times \frac{C}{N+1} \times H \times W}$, where $B$, $\frac{C}{N+1}$, $H$, and $W$ represent the batch size, channel dimension, height, and width, respectively, is initially fed into the Layer Operator (LO) $1$, resulting in a new head $A_2$. 
The channel dimension of the output head $A_2$ increases from $\frac{C}{N+1}$ to $\frac{2}{N+1}C$.
Subsequently, the output head $A_2$ is recursively fed into the LO $n$, illustrated in \cref{fig:Recursive-Convolution} (right), according to the layer number $n$ ($n\in{[2, N]}$), with its channel dimension increasing from $\frac{n}{N+1}C$ to $\frac{n+1}{N+1}C$.
The remaining $N$ input heads, $H_1$, ..., $H_n$, ..., $H_N$ are sequentially fed into the LO $n$ according to the layer number $n$ ($n\in{[1, N]}$) to interact with the corresponding input head $A_n$. 
Each head is initially projected using $1\times{1}$ convolution before feeding in the LO to enhance feature diversity.
In the RFA, the output channels of heads $A_n$ follow a pyramidal increment, reducing parameters and FLOPs compared to the standard direct-in, direct-out construction.
Increasing the layer $N$ allows for higher input image resolution, potentially providing a more effective alternative to training on low-resolution images followed by fine-tuning at high resolutions.



%

\subsection{Layer Operator}


To effectively expand receptive field and assign discriminative impact on receptive field, 
we introduce the Layer Operator (LO). This operator is designed from the perspective of receptive field and serves as the core operator in the RFA, illustrated in \cref{fig:Recursive-Convolution} (right). 
This technique can construct a two-layer AGD of receptive field.
Specifically, for layer number $n$, the three distinct inputs for LO $n$ are $a_{n,1}$, $a_{n,2}$, and $h_n$, projected by three individual $1\times{1}$ convolutions as illustrated in \cref{fig:Recursive-Convolution} (left). 
The LO is generated by interacting two components, the Discriminator (Dis) and the Amplifier (Amp), as illustrated in \cref{fig:Recursive-Convolution} (right).
For the Amp, we conduct an element-wise multiplication between $a_{n,2}$ and output features derived from a GELU activation and a depth-wise large-kernel $K\times{K}$ convolution applied to $a_{n,1}$.
For each output pixel in the Amp module, the corresponding pixels within the relevant $K \times K$ receptive field are multiplied by the pixel value at the same spatial location in $a_{n,2}$. This operation expands the receptive field and amplifies the impact of pixels on the receptive field.
Additionally, for the Dis, we incorporate features from depth-wise $K \times K$ and $k \times k$ convolutions.
This introduces impact from small-scale new pixels for the large $K \times K$ receptive field, establishing a two-layer discriminative AGD.
The $1\times1$ convolutions facilitate information interaction among channels and change the channel dimension for the compatibility of features. 
The outputs of the Amp and the Dis are concatenated, resulting in the final output $A_{n+1}$ with a two-layer AGD of receptive field and increased channels for the subsequent layer.

\begin{figure}[t]
	\centering
	\includegraphics[width=\linewidth]{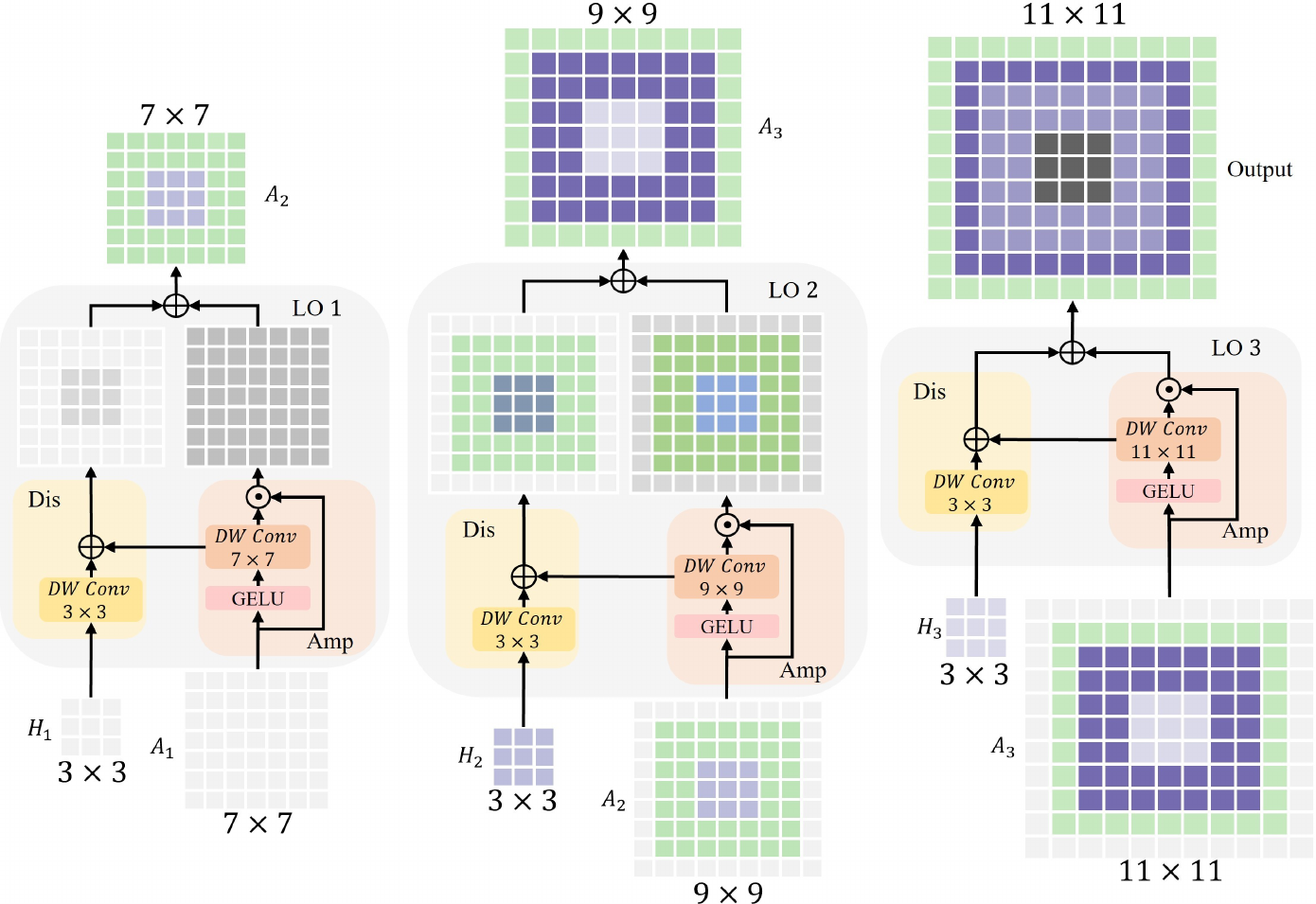}
	
	\caption{\textbf{\bfseries{Receptive Field Flow of the Three-layer Receptive Field Aggregator.}}}
	\label{fig:receptive field flow}
\end{figure}

\subsection{Three-layer RFA for UniConvNet}
\label{Three-layer RFA for UniConvNet}

The layer number $N$ varies with the resolution of the input image. 
In this work, for the input image size of $224\times{224}$, we construct a three-layer RFA using a layer $N=3$. 
The progressive large-scale kernel size $K$ in the LO is calculated as $K=2n+5(n\in{[1, N]})$. 
The small-scale kernel size is $k=3$.
Thus, the RFA achieves a four-layer AGD for receptive field within a shadow module using progressive large-kernel convolutions $7\times{7}$, $9\times{9}$ and $11\times{11}$, which is an optimal configuration for $224\times{224}$ images. We do an ablation study in \cref{sec:Ablation Study}.

\begin{figure}[t]
	\centering
	\includegraphics[width=\linewidth]{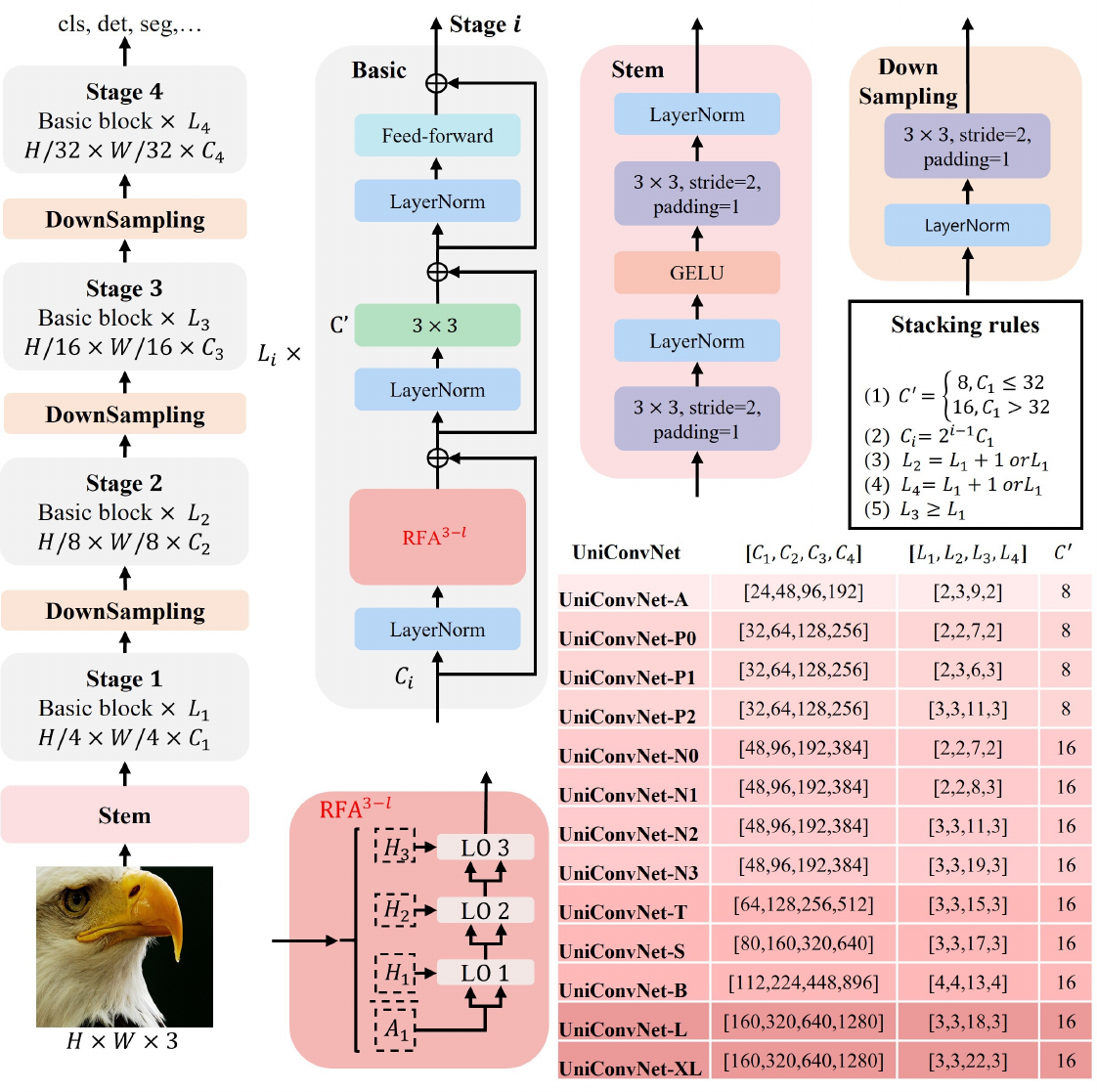}
	
	\caption{\textbf{\bfseries{Overall architecture of UniConvNet.}}}
	\label{fig:Architecture}
\end{figure}

The smallest kernel size of $7\times7$ provides a significantly larger receptive field than $3\times3$ and $5\times5$ convolutions. 
The largest convolution kernel size of $11\times11$ can maintain the $14\times14$ features with padding size 5 in stage 3, the main stage of the feature extraction in our models, depicted in \cref{fig:Architecture}, of our model to have at most quarter pixels of the feature at the corner for avoiding the frequent overlapping of the center pixels during the convolution progress.
The efficiency and effectiveness are justified and analyzed in \cref{Efficiency and Effectiveness of Three-layer RFA}.

\subsection{Receptive Field Flow of Three-layer RFA}
\label{Receptive Field Flow of Three-layer RFA}

We sketch the receptive field flow of the proposed three-layer RFA in \cref{fig:receptive field flow}.
In LO 1, the receptive field scale is expanded and the impacts of pixels on $7\times{7}$ receptive field is amplified by the Amp module. 
The Dis module provide the impacts of small-scale new pixels, closer to the position of output pixel, for large receptive field generated by $7\times{7}$ convolution to build a discriminative receptive field. 
The output combines the two receptive fields to build a two-layer AGD by directly assigning impacts of different scales on the receptive field.
Sequentially, each LO expands and amplifies the receptive field of the previous LO by Amp and provides a discriminative receptive field for adding impacts of small-scale new pixels. 
The final receptive field of three-layer RFA results in a four-layer receptive field following an AGD.
The ERF can be expanded and the AGD of ERF can be maintained through the stack of many RFA modules, as illustrated in \cref{fig:intro} (D).

\subsection{UniConvNet Model}


The Three-layer RFA can be integrated into any model as a plug-and-play module, effectively replacing each convolution in ConvNet architectures.
For better performance, we directly integrate the Three-layer RFA module into the state-of-the-art CNN-based model, InternImage \cite{DCNV3}. Specifically, the convolution used in the $3\times{3}$ convolution residual component is the DCNV3 in the InternImage \cite{DCNV3} and we remove the softmax normalization in DCNV3 as DCNV4 did because the optimizations for memory on DCNV4 in the Efficient Deformable ConvNets \cite{DCNV4} is incompatible with the other CNN-based models. We develop a new CNN-based backbone model called UniConvNet using various stacking strategies for its variants. 
%
%
%
%
%
The overall architecture and stacking rules are illustrated in \cref{fig:Architecture}. 
We present integral blocks of the model and provide details on the embedded blocks in \cref{Illustration of UniConvNet Block}.

\begin{table}
	\belowrulesep=0pt
	\aboverulesep=0pt
	\centering
	\resizebox{0.9\linewidth}{!}
	{%
			\begin{tabular}{@{}lcccccc@{}}
					\textbf{\bfseries{Model}} & 
					\textbf{\bfseries{Type}}& 
					\textbf{\bfseries{Scale}}& 
					\textbf{\bfseries{\#Params}}& 
					\textbf{\bfseries{FLOPs}}& 
					\textbf{\bfseries{Acc(\%)}}& 
					\textbf{\bfseries{Publication}}
					\\
					\midrule
					EdgeViT-XXS \cite{EdgeViT} & 
					T & 
					256\textsuperscript{2} & 
					4.1M & 
					0.577G & 
					74.4 & 
					ECCV'22
					\\
					FastViT-T8 \cite{FastViT} & 
					T & 
					256\textsuperscript{2} & 
					3.6M & 
					0.7G & 
					75.6 & 
					ICCV'23
					\\
					PVTv2-B0 \cite{PVTv2} & 
					T & 
					224\textsuperscript{2} & 
					3.7M & 
					0.572G & 
					70.5 & 
					CVM'22
					\\
					tiny-MOAT-0 \cite{MOAT} & 
					T & 
					224\textsuperscript{2} & 
					3.4M & 
					0.8G & 
					75.5 & 
					ICLR'23
					\\
					\rowcolor{yellow!10}
					UniRepLKNet-A \cite{UniRepLKNet} & 
					C & 
					224\textsuperscript{2} & 
					4.4M & 
					0.6G & 
					77.0 & 
					CVPR'24
					\\
					\rowcolor{yellow!10}
					StarNet-S2 \cite{StarNet} & 
					C & 
					224\textsuperscript{2} & 
					3.7M & 
					0.547G & 
					74.8 & 
					CVPR'24
					\\
					\rowcolor{blue!12}
					\textbf{\bfseries{UniConvNet-A(ours)}} & 
					\textbf{\bfseries{C}} & 
					\textbf{\bfseries{224\textsuperscript{2}}} & 
					\textbf{\bfseries{3.4M}} & 
					\textbf{\bfseries{0.589G}} & 
					\textbf{\bfseries{77.0}} & 
					\textbf{\bfseries{-}}
					\\
					\midrule
					EdgeNeXt-S \cite{EdgeNeXt} & 
					T & 
					256\textsuperscript{2} & 
					5.6M & 
					0.965G & 
					78.8 & 
					ECCVW'22
					\\
					MobileViTv1-S \cite{MobileViTv1} & 
					T & 
					256\textsuperscript{2} & 
					5.6M & 
					2.01G & 
					78.4 & 
					ICLR'22
					\\
					tiny-MOAT-1 \cite{MOAT} & 
					T & 
					224\textsuperscript{2} & 
					5.1M & 
					1.2G & 
					78.3 & 
					ICLR'23
					\\
					EMO-5M \cite{EMO} & 
					T & 
					224\textsuperscript{2} & 
					5.1M & 
					0.903G & 
					78.4 & 
					ICCV'23
					\\
					\rowcolor{yellow!10}
					StarNet-S3 \cite{StarNet} & 
					C & 
					224\textsuperscript{2} & 
					5.8M & 
					0.757G & 
					77.3 & 
					CVPR'24
					\\
					\rowcolor{yellow!10}
					RepViT-M0.9 \cite{RepViT} & 
					C & 
					224\textsuperscript{2} & 
					5.1M & 
					0.8G & 
					77.4 & 
					CVPR'24
					\\
					\rowcolor{yellow!10}
					DCNV4 \cite{DCNV4} & 
					C & 
					224\textsuperscript{2} & 
					5.3M & 
					0.805G & 
					78.5 & 
					CVPR'24
					\\
					\rowcolor{blue!12}
					\textbf{\bfseries{UniConvNet-P0(ours)}} & 
					\textbf{\bfseries{C}} & 
					\textbf{\bfseries{224\textsuperscript{2}}} & 
					\textbf{\bfseries{5.2M}} & 
					\textbf{\bfseries{0.832G}} & 
					\textbf{\bfseries{79.1}} & 
					\textbf{\bfseries{-}}
					\\
					\midrule
					EdgeViT-XS \cite{EdgeViT} & 
					T & 
					256\textsuperscript{2} & 
					6.7M & 
					1.136G & 
					77.5 & 
					ECCV'22
					\\
					FastViT-T12 \cite{FastViT} & 
					T & 
					256\textsuperscript{2} & 
					6.8M & 
					1.4G & 
					79.1 & 
					ICCV'23
					\\
					SMT-M \cite{SMT} & 
					T & 
					224\textsuperscript{2} & 
					6.5M & 
					1.3G & 
					78.4 & 
					ICCV'23
					\\
					EMO-6M \cite{EMO} & 
					T & 
					224\textsuperscript{2} & 
					6.1M & 
					0.961G & 
					79.0 & 
					ICCV'23
					\\
					\rowcolor{yellow!10}
					StarNet-S4 \cite{StarNet} & 
					C & 
					224\textsuperscript{2} & 
					7.5M & 
					1.075G & 
					78.4 & 
					CVPR'24
					\\
					\rowcolor{yellow!10}
					RepViT-M1.0 \cite{RepViT} & 
					C & 
					224\textsuperscript{2} & 
					6.8M & 
					1.1G & 
					78.6 & 
					CVPR'24
					\\
					\rowcolor{yellow!10}
					UniRepLKNet-F \cite{UniRepLKNet} & 
					C & 
					224\textsuperscript{2} & 
					6.2M & 
					0.9G & 
					78.6 & 
					CVPR'24
					\\
					\rowcolor{blue!12}
					\textbf{\bfseries{UniConvNet-P1(ours)}} & 
					\textbf{\bfseries{C}} & 
					\textbf{\bfseries{224\textsuperscript{2}}} & 
					\textbf{\bfseries{6.1M}} & 
					\textbf{\bfseries{0.895G}} & 
					\textbf{\bfseries{79.6}} & 
					\textbf{\bfseries{-}}
					\\
					\midrule
					FastViT-S12 \cite{FastViT} & 
					T & 
					256\textsuperscript{2} & 
					8.8M & 
					1.8G & 
					79.8 & 
					ICCV'23
					\\
					\rowcolor{yellow!10}
					EfficientNet-B2 \cite{EfficientNet} & 
					C & 
					288\textsuperscript{2} & 
					9.2M & 
					1.0G & 
					80.1 & 
					ICCV'19
					\\
					\rowcolor{yellow!10}
					RepViT-M1.1 \cite{RepViT} & 
					C & 
					224\textsuperscript{2} & 
					8.2M & 
					1.3G & 
					79.4 & 
					CVPR'24
					\\
					\rowcolor{blue!12}
					\textbf{\bfseries{UniConvNet-P2(ours)}} & 
					\textbf{\bfseries{C}} & 
					\textbf{\bfseries{224\textsuperscript{2}}} & 
					\textbf{\bfseries{7.6M}} & 
					\textbf{\bfseries{1.25G}} & 
					\textbf{\bfseries{80.5}} & 
					\textbf{\bfseries{-}}
					\\
					\midrule
					Shunted-T \cite{ShuntedSA} & 
					T & 
					256\textsuperscript{2} & 
					11.5M & 
					2.1G & 
					79.8 & 
					CVPR'22
					\\
					EdgeViT-S \cite{EdgeViT} & 
					T & 
					256\textsuperscript{2} & 
					11.1M & 
					1.9G & 
					81.0 & 
					ECCV'22
					\\
					FastViT-SA12 \cite{FastViT} & 
					T & 
					256\textsuperscript{2} & 
					10.9M & 
					1.9G & 
					80.6 & 
					ICCV'23
					\\
					EfficientFormer-L1 \cite{Efficientformer} & 
					T & 
					224\textsuperscript{2} & 
					12.3M & 
					1.3G & 
					79.1 & 
					NeurIPS'22
					\\
					MpViT-XS \cite{MPViT} & 
					T & 
					224\textsuperscript{2} & 
					10.5M & 
					2.9G & 
					80.9 & 
					CVPR'22
					\\
					\rowcolor{yellow!10}
					EfficientNet-B3 \cite{EfficientNet} & 
					C & 
					300\textsuperscript{2} & 
					12.0M & 
					1.8G & 
					81.6 & 
					ICCV'19
					\\
					\rowcolor{blue!12}
					\textbf{\bfseries{UniConvNet-N0(ours)}} & 
					\textbf{\bfseries{C}} & 
					\textbf{\bfseries{224\textsuperscript{2}}} & 
					\textbf{\bfseries{10.2M}} & 
					\textbf{\bfseries{1.65G}} & 
					\textbf{\bfseries{81.6}} & 
					\textbf{\bfseries{-}}
					\\
					\midrule
					PVTv1-Tiny \cite{Pvtv1} & 
					T & 
					224\textsuperscript{2} & 
					13.2M & 
					1.9G & 
					75.1 & 
					ICCV'21
					\\
					\rowcolor{yellow!10}
					RepViT-M1.5 \cite{RepViT} & 
					C & 
					224\textsuperscript{2} & 
					14.0M & 
					2.3G & 
					81.2 & 
					CVPR'24
					\\
					\rowcolor{blue!12}
					\textbf{\bfseries{UniConvNet-N1(ours)}} & 
					\textbf{\bfseries{C}} & 
					\textbf{\bfseries{224\textsuperscript{2}}} & 
					\textbf{\bfseries{13.1M}} & 
					\textbf{\bfseries{1.88G}} & 
					\textbf{\bfseries{82.2}} & 
					\textbf{\bfseries{-}}
					\\
					\midrule
					FastViT-SA24 \cite{FastViT} & 
					T & 
					256\textsuperscript{2} & 
					20.6M & 
					3.8G & 
					82.6 & 
					ICCV'23
					\\
					Dilate-T \cite{Dilateformer} & 
					T & 
					224\textsuperscript{2} & 
					17.0M & 
					3.2G & 
					82.1 & 
					TMM'23
					\\
					\rowcolor{yellow!10}
					RepViT-M2.3 \cite{RepViT} & 
					C & 
					224\textsuperscript{2} & 
					22.9M & 
					4.5G & 
					82.5 & 
					CVPR'24
					\\
					\rowcolor{yellow!10}
					UniRepLKNet-N \cite{UniRepLKNet} & 
					C & 
					224\textsuperscript{2} & 
					18.3M & 
					2.8G & 
					81.6 & 
					CVPR'24
					\\
					\rowcolor{blue!12}
					\textbf{\bfseries{UniConvNet-N2(ours)}} & 
					\textbf{\bfseries{C}} & 
					\textbf{\bfseries{224\textsuperscript{2}}} & 
					\textbf{\bfseries{15.0M}} & 
					\textbf{\bfseries{2.47G}} & 
					\textbf{\bfseries{82.7}} & 
					\textbf{\bfseries{-}}
					\\
					\midrule
					XCiT-S12/16 \cite{XCiT} & 
					T & 
					224\textsuperscript{2} & 
					26.0M & 
					4.8G & 
					82.0 & 
					NeurIPS'21
					\\
					ViTAE-S \cite{Vitae} & 
					T & 
					224\textsuperscript{2} & 
					23.6M & 
					5.6G & 
					82.0 & 
					NeurIPS'21
					\\
					CoAtNet-0 \cite{CoAtNet} & 
					T & 
					224\textsuperscript{2} & 
					25.0M & 
					4.0G & 
					81.6 & 
					NeurIPS'21
					\\
					PVTv2-B2-Li \cite{PVTv2} & 
					T & 
					224\textsuperscript{2} & 
					22.6M & 
					3.9G & 
					82.1 & 
					CVM'22
					\\
					\rowcolor{yellow!10}
					HorNet-T \cite{hornet} & 
					C & 
					224\textsuperscript{2} & 
					23.0M & 
					4.0G & 
					83.0 & 
					NeurIPS'22
					\\
					\rowcolor{blue!12}
					\textbf{\bfseries{UniConvNet-N3(ours)}} & 
					\textbf{\bfseries{C}} & 
					\textbf{\bfseries{224\textsuperscript{2}}} & 
					\textbf{\bfseries{19.7M}} & 
					\textbf{\bfseries{3.37G}} & 
					\textbf{\bfseries{83.2}} & 
					\textbf{\bfseries{-}}
					\\
				\end{tabular}
		}
	\caption{\textbf{\bfseries{Image classification performance on the ImageNet validation set for lightweight variants.}} “Type” refers to model type, where “C” and “T” denote pure CNN and the model using the Transformer for its architecture, respectively. “Scale” is the input image scale. “Acc” is the TOP-1 accuracy.}
	\label{tab:table-classification-lightweight}
\end{table}

\begin{table}
	\belowrulesep=0pt
	\aboverulesep=0pt
	\centering
	\resizebox{0.9\linewidth}{!}
	{%
		\begin{tabular}{@{}lcccccc@{}}
			\textbf{\bfseries{Model}} & 
			\textbf{\bfseries{Type}}& 
			\textbf{\bfseries{Scale}}& 
			\textbf{\bfseries{\#Params}}& 
			\textbf{\bfseries{FLOPs}}& 
			\textbf{\bfseries{Acc(\%)}}& 
			\textbf{\bfseries{Publication}}
			\\
			\midrule
			CoAtNet-1 \cite{CoAtNet} & 
			T & 
			224\textsuperscript{2} & 
			42.0M & 
			8.0G & 
			83.3 & 
			NeurIPS'21
			\\
			EfficientFormer-L3 \cite{Efficientformer} & 
			T & 
			224\textsuperscript{2} & 
			31.3M & 
			3.9G & 
			82.4 & 
			NeurIPS'22
			\\
			Swin-T \cite{Swin-transformer} & 
			T & 
			224\textsuperscript{2} & 
			29.0M & 
			5.0G & 
			81.3 & 
			ICCV'21
			\\
			Focal-T \cite{Focal} & 
			T & 
			224\textsuperscript{2} & 
			29.0M & 
			5.0G & 
			82.2 & 
			NeurIPS'21
			\\
			CrossViT-18 \cite{CrossViT} & 
			T & 
			224\textsuperscript{2} & 
			28.2M & 
			6.1G & 
			82.3 & 
			ICCV'21
			\\
			\rowcolor{yellow!10}
			UniRepLKNet-T \cite{UniRepLKNet} & 
			C & 
			224\textsuperscript{2} & 
			31.0M & 
			4.9G & 
			83.2 & 
			CVPR'24
			\\
			\rowcolor{yellow!10}
			SLaK-T \cite{SLaK} & 
			C & 
			224\textsuperscript{2} & 
			30.0M & 
			5.0G & 
			82.5 & 
			ICLR'23
			\\
			\rowcolor{yellow!10}
			InternImage-T \cite{DCNV3} & 
			C & 
			224\textsuperscript{2} & 
			30.0M & 
			5.0G & 
			83.5 & 
			CVPR'23
			\\
			\rowcolor{yellow!10}
			FlashInternImage-T \cite{DCNV4} & 
			C & 
			224\textsuperscript{2} & 
			30.0M & 
			- & 
			83.6 & 
			CVPR'24
			\\
			\rowcolor{yellow!10}
			PeLK-T \cite{pelk} & 
			C & 
			224\textsuperscript{2} & 
			29.0M & 
			5.6G & 
			82.6 & 
			CVPR'24
			\\
			\rowcolor{yellow!10}
			ConvNeXt-T \cite{ConvNeXt} & 
			C & 
			224\textsuperscript{2} & 
			29.0M & 
			5.0G & 
			82.1 & 
			CVPR'22
			\\
			\rowcolor{yellow!10}
			MogaNet-S \cite{moganet} & 
			C & 
			224\textsuperscript{2} & 
			25.0M & 
			5.0G & 
			83.4 & 
			ICLR'24
			\\
			WTConvNeXt-T \cite{WTConvNeXt} & 
			- & 
			224\textsuperscript{2} & 
			30.0M & 
			4.5G & 
			82.5 & 
			ECCV'24
			\\
			\rowcolor{blue!10}
			\textbf{\bfseries{UniConvNet-T(ours)}} & 
			\textbf{\bfseries{C}} & 
			\textbf{\bfseries{224\textsuperscript{2}}} & 
			\textbf{\bfseries{30.3M}} & 
			\textbf{\bfseries{5.1G}} & 
			\textbf{\bfseries{84.2}} & 
			\textbf{\bfseries{-}}
			\\
			\midrule
			SwinV2-S/8 \cite{Swin-transformer-v2} & 
			T & 
			256\textsuperscript{2} & 
			50.0M & 
			12.0G & 
			83.7 & 
			CVPR'22
			\\
			CoAtNet-2 \cite{CoAtNet} & 
			T & 
			224\textsuperscript{2} & 
			75.0M & 
			16.0G & 
			84.1 & 
			NeurIPS'21
			\\
			PVTv2-B4 \cite{PVTv2} & 
			T & 
			224\textsuperscript{2} & 
			63.0M & 
			10.0G & 
			83.6 & 
			CVM'22
			\\
			Swin-S \cite{Swin-transformer} & 
			T & 
			224\textsuperscript{2} & 
			50.0M & 
			9.0G & 
			83.0 & 
			ICCV'21
			\\		
			\rowcolor{yellow!10}
			RepLKNet-31B \cite{RepLKNet} & 
			C & 
			224\textsuperscript{2} & 
			79.0M & 
			15.0G & 
			83.5 & 
			CVPR'22
			\\
			\rowcolor{yellow!10}
			PeLK-S \cite{pelk} & 
			C & 
			224\textsuperscript{2} & 
			50.0M & 
			10.7G & 
			83.9 & 
			CVPR'24
			\\
			\rowcolor{yellow!10}
			SLaK-S \cite{SLaK} & 
			C & 
			224\textsuperscript{2} & 
			55.0M & 
			10.0G & 
			83.8 & 
			ICLR'23
			\\
			\rowcolor{yellow!10}
			ConvNeXt-S \cite{ConvNeXt} & 
			C & 
			224\textsuperscript{2} & 
			50.0M & 
			9.0G & 
			83.1 & 
			CVPR'22
			\\
			\rowcolor{yellow!10}
			HorNet-S \cite{hornet} & 
			C & 
			224\textsuperscript{2} & 
			50.0M & 
			9.0G & 
			84.0 & 
			NeurIPS'22
			\\
			\rowcolor{yellow!10}
			InternImage-S \cite{DCNV3} & 
			C & 
			224\textsuperscript{2} & 
			50.0M & 
			8.0G & 
			84.2 & 
			CVPR'23
			\\
			\rowcolor{yellow!10}
			FlashInternImage-S \cite{DCNV4} & 
			C & 
			224\textsuperscript{2} & 
			50.0M & 
			- & 
			84.4 & 
			CVPR'24
			\\
			WTConvNeXt-S \cite{WTConvNeXt} & 
			- & 
			224\textsuperscript{2} & 
			54.0M & 
			8.8G & 
			83.6 & 
			ECCV'24
			\\
			\rowcolor{blue!10}
			\textbf{\bfseries{UniConvNet-S(ours)}} & 
			\textbf{\bfseries{C}} & 
			\textbf{\bfseries{224\textsuperscript{2}}} & 
			\textbf{\bfseries{50.0M}} & 
			\textbf{\bfseries{8.48G}} & 
			\textbf{\bfseries{84.5}} & 
			\textbf{\bfseries{-}}
			\\
			\midrule
			SwinV2-B/8 \cite{Swin-transformer-v2} & 
			T & 
			256\textsuperscript{2} & 
			88.0M & 
			20.0G & 
			84.2 & 
			CVPR'22
			\\
			Swin-B \cite{Swin-transformer} & 
			T & 
			224\textsuperscript{2} & 
			88.0M & 
			15.0G & 
			83.5 & 
			ICCV'21
			\\
			PVTv2-B5 \cite{PVTv2} & 
			T & 
			224\textsuperscript{2} & 
			82.0M & 
			12.0G & 
			83.8 & 
			CVM'22
			\\
			\rowcolor{yellow!10}
			ConvNeXt-B \cite{ConvNeXt} & 
			C & 
			224\textsuperscript{2} & 
			88.0M & 
			15.0G & 
			83.8 & 
			CVPR'22
			\\
			\rowcolor{yellow!10}
			InternImage-B \cite{DCNV3} & 
			C & 
			224\textsuperscript{2} & 
			97.0M & 
			16.0G & 
			84.9 & 
			CVPR'23
			\\
			\rowcolor{yellow!10}
			FlashInternImage-B \cite{DCNV4} & 
			C & 
			224\textsuperscript{2} & 
			97.0M & 
			- & 
			84.9 & 
			CVPR'24
			\\
			\rowcolor{yellow!10}
			SLaK-B \cite{SLaK} & 
			C & 
			224\textsuperscript{2} & 
			95.0M & 
			17.0G & 
			84.0 & 
			ICLR'23
			\\
			\rowcolor{yellow!10}
			PeLK-B \cite{pelk} & 
			C & 
			224\textsuperscript{2} & 
			89.0M & 
			18.3G & 
			84.2 & 
			CVPR'24
			\\
			\rowcolor{yellow!10}
			HorNet-B \cite{hornet} & 
			C & 
			224\textsuperscript{2} & 
			88.0M & 
			16.0G & 
			84.3 & 
			NeurIPS'22
			\\
			WTConvNeXt-B \cite{WTConvNeXt} & 
			- & 
			224\textsuperscript{2} & 
			93.0M & 
			15.5G & 
			84.1 & 
			ECCV'24
			\\
			\rowcolor{blue!10}
\textbf{\bfseries{UniConvNet-B(ours)}} & 
\textbf{\bfseries{C}} & 
\textbf{\bfseries{224\textsuperscript{2}}} & 
\textbf{\bfseries{97.6M}} & 
\textbf{\bfseries{15.9G}} & 
\textbf{\bfseries{85.0}} & 
\textbf{\bfseries{-}}
\\
\midrule
Swin-B \cite{Swin-transformer} & 
T & 
384\textsuperscript{2} & 
88.0M & 
47.1G & 
84.5 & 
ICCV'21
\\
\rowcolor{yellow!10}
ConvNeXt-B \cite{ConvNeXt} & 
C & 
384\textsuperscript{2} & 
89.0M & 
45.0G & 
85.1 & 
CVPR'22
\\
\rowcolor{blue!10}
\textbf{\bfseries{UniConvNet-T(ours)}} & 
\textbf{\bfseries{C}} & 
\textbf{\bfseries{384\textsuperscript{2}}} & 
\textbf{\bfseries{30.3M}} & 
\textbf{\bfseries{15.0G}} & 
\textbf{\bfseries{85.4}} & 
\textbf{\bfseries{-}}
\\
\rowcolor{yellow!10}
ConvNeXt-L \cite{ConvNeXt} & 
C & 
384\textsuperscript{2} & 
198.0M & 
101.0G & 
85.5 & 
CVPR'22
\\
\rowcolor{yellow!10}
PeLK-B \cite{pelk} & 
C & 
384\textsuperscript{2} & 
89.0M & 
54.0G & 
85.6 & 
CVPR'24
\\
\rowcolor{blue!10}
\textbf{\bfseries{UniConvNet-S(ours)}} & 
\textbf{\bfseries{C}} & 
\textbf{\bfseries{384\textsuperscript{2}}} & 
\textbf{\bfseries{50.0M}} & 
\textbf{\bfseries{24.9G}} & 
\textbf{\bfseries{85.7}} & 
\textbf{\bfseries{-}}
\\
\rowcolor{yellow!10}
PeLK-B-101 \cite{pelk} & 
C & 
384\textsuperscript{2} & 
90.0M & 
68.3G & 
85.8 & 
CVPR'24
\\
\rowcolor{blue!10}
\textbf{\bfseries{UniConvNet-B(ours)}} & 
\textbf{\bfseries{C}} & 
\textbf{\bfseries{384\textsuperscript{2}}} & 
\textbf{\bfseries{97.6M}} & 
\textbf{\bfseries{46.6G}} & 
\textbf{\bfseries{85.9}} & 
\textbf{\bfseries{-}}
\\
\midrule
SwinV2-L/24\dag \cite{Swin-transformer-v2} & 
T & 
384\textsuperscript{2} & 
197.0M & 
115.0G & 
87.6 & 
CVPR'22
\\
Swin-L\dag \cite{Swin-transformer} & 
T & 
384\textsuperscript{2} & 
197.0M & 
104.0G & 
87.3 & 
ICCV'21
\\
MOAT-3\dag \cite{MOAT} & 
T & 
384\textsuperscript{2} & 
190.0M & 
141.2G & 
88.2 & 
ICLR'23
\\

\rowcolor{yellow!10}
HorNet-L\dag \cite{hornet} & 
C & 
384\textsuperscript{2} & 
202.0M & 
102.0G & 
87.7 & 
NeurIPS'22
\\
\rowcolor{yellow!10}
ConvNeXt-L\dag \cite{ConvNeXt} & 
C & 
384\textsuperscript{2} & 
198.0M & 
101.0G & 
87.5 & 
CVPR'22
\\
\rowcolor{yellow!10}
MogaNet-XL \cite{moganet} & 
C & 
384\textsuperscript{2} & 
181.0M & 
102.0G & 
87.8 & 
ICLR'24
\\
\rowcolor{yellow!10}
RepLKNet-31L\dag \cite{RepLKNet} & 
C & 
384\textsuperscript{2} & 
172.0M & 
96.0G & 
86.6 & 
CVPR'22
\\
\rowcolor{blue!10}
\textbf{\bfseries{UniConvNet-L\dag(ours)}} & 
\textbf{\bfseries{C}} & 
\textbf{\bfseries{384\textsuperscript{2}}} & 
\textbf{\bfseries{201.8M}} & 
\textbf{\bfseries{100.1G}} & 
\textbf{\bfseries{88.2}} & 
\textbf{\bfseries{-}}
\\
\midrule
CoAtNet-4\dag \cite{CoAtNet} & 
T & 
384\textsuperscript{2} & 
275.0M & 
190.0G & 
87.9 & 
NeurIPS'21
\\
\rowcolor{yellow!10}
ConvNeXt-XL\dag \cite{ConvNeXt} & 
C & 
384\textsuperscript{2} & 
350.0M & 
179.0G & 
87.7 & 
CVPR'22
\\
\rowcolor{yellow!10}
InternImage-XL\dag \cite{DCNV3} & 
C & 
384\textsuperscript{2} & 
335.0M & 
163.0G & 
88.0 & 
CVPR'23
\\
\rowcolor{yellow!10}
FlashInternImage-L\dag \cite{DCNV4} & 
C & 
384\textsuperscript{2} & 
224.0M & 
- & 
88.1 & 
CVPR'24
\\
\rowcolor{yellow!10}
InternImage-L\dag \cite{DCNV3} & 
C & 
384\textsuperscript{2} & 
223.0M & 
108.0G & 
87.7 & 
CVPR'23
\\
\rowcolor{blue!10}
\textbf{\bfseries{UniConvNet-XL\dag(ours)}} & 
\textbf{\bfseries{C}} & 
\textbf{\bfseries{384\textsuperscript{2}}} & 
\textbf{\bfseries{226.7M}} & 
\textbf{\bfseries{115.2G}} & 
\textbf{\bfseries{88.4}} & 
\textbf{\bfseries{-}}
\\

\end{tabular}
}
\caption{\textbf{\bfseries{Image classification performance on the ImageNet validation set for scaled-up variants.}} “Type” refers to model type, where “C” and “T” denote pure CNN and the model using the Transformer for its architecture, respectively. “Scale” is the input image scale. “Acc” is the TOP-1 accuracy. “\dag”  indicates the model is pre-trained on ImageNet-22K\cite{ImageNet}.}
\label{tab:table-classification-scaled-up}
\end{table}


\section{Experiments}

We develop different UniConvNet variants to match the complexities of various contemporary models, including state-of-the-art lightweight networks \cite{EMO,FastViT,StarNet,RepViT} and large-scale networks \cite{ConvNeXt,Swin-transformer,MOAT,UniRepLKNet,DCNV3,DCNV4}.
We evaluate the performance of UniConvNet variants and compare them to leading CNNs and ViTs across representative vision tasks, including image classification, object detection, and instance and semantic segmentation.

\subsection{Image Classification}


For a fair comparison, in line with common practices \cite{ ConvNeXt, DCNV4}, UniConvNet-A/P0/P1/P2/N0/N1/N2/N3/T/S/B are trained on ImageNet-1K for 300 epochs, while UniConvNet-L is first trained on ImageNet-22K for 90 epochs and then fine-tuned on ImageNet-1K for 20 epochs.
Detailed ImageNet-1K/22K training settings, ImageNet-1K fine-tune settings and training recipes for different variants are presented in \cref{ImageNet-1K/22K Training Settings},  \ref{ImageNet-1K Fine-tune Settings} and \ref{Selection-Training-recipe}, respectively.


As is shown in \cref{tab:table-classification-lightweight} and \cref{tab:table-classification-scaled-up}, our proposed model variants demonstrate significant improvements over state-of-the-art models, effectively bridging the gap between lightweight and large-scale models. Existing models either exhibit inferior performance in lightweight scenarios or fail to achieve adequate accuracy when scaled up. 



%

%

\begin{table}
	\belowrulesep=0pt
	\aboverulesep=0pt
	\centering
	\resizebox{\linewidth}{!}
	{%
			\begin{tabular}{@{}lccccccccc@{}}
					\multicolumn{10}{c}{\textbf{\bfseries{heavy RetinaNet}}}
					\\
					\textbf{\bfseries{Backbone}} & 
					\textbf{\bfseries{Type}}& 
					\textbf{\bfseries{\#Params}}& 
					\textbf{\bfseries{FLOPs}}&  
					\textbf{$\boldsymbol{mAP}$} &
					\textbf{$\boldsymbol{mAP_{50}}$} &
					\textbf{$\boldsymbol{mAP_{75}}$} &
					\textbf{$\boldsymbol{mAP_{S}}$} &
					\textbf{$\boldsymbol{mAP_{M}}$} &
					\textbf{$\boldsymbol{mAP_{L}}$}
					\\
					\midrule
					EdgeViT-XXS \cite{EdgeViT} & 
					T & 
					13.1M & 
					 - &
					38.7 & 
					59.0 & 
					41.0 & 
					22.4 & 
					42.0 & 
					51.6 
					\\
					EMO-5M \cite{EMO} & 
					T & 
					14.4M & 
					 - &
					38.9 & 
					59.8 & 
					41.0 & 
					23.8 & 
					42.2 & 
					51.7 
					\\
					\rowcolor{blue!10}
					\textbf{\bfseries{UniConvNet-A(ours)}} & 
					\textbf{\bfseries{C}} & 
					\textbf{\bfseries{12.6M}} &
					\textbf{\bfseries{16.3G}} &
					\textbf{\bfseries{40.0}} & 
					\textbf{\bfseries{60.4}} & 
					\textbf{\bfseries{42.9}} & 
					\textbf{\bfseries{23.3}} & 
					\textbf{\bfseries{44.2}} &
					\textbf{\bfseries{53.2}} 
					\\
					\midrule
					Mobile-Former-151M \cite{Mobile-former} & 
					T & 
					14.4M & 
					 - &
					34.2 & 
					53.4 & 
					36.0 & 
					19.9 & 
					36.8 & 
					45.3 
					\\
					Mobile-Former-214M \cite{Mobile-former} & 
					T & 
					15.2M & 
					 - &
					35.8 & 
					55.4 & 
					38.0 & 
					21.8 & 
					38.5 & 
					46.8 
					\\
					Mobile-Former-294M \cite{Mobile-former} & 
					T & 
					16.1M & 
					- &
					36.6 & 
					56.6 & 
					38.6 & 
					21.9 & 
					39.5 & 
					47.9 
					\\
					EdgeViT-XS \cite{EdgeViT} & 
					T & 
					16.3M & 
					 - &
					40.6 & 
					61.3 & 
					43.3 & 
					25.2 & 
					43.9 & 
					54.6 
					\\
					\rowcolor{blue!10}
					\textbf{\bfseries{UniConvNet-P0(ours)}} & 
					\textbf{\bfseries{C}} & 
					\textbf{\bfseries{14.4M}} &
					\textbf{\bfseries{16.8G}} &
					\textbf{\bfseries{41.1}} & 
					\textbf{\bfseries{61.4}} & 
					\textbf{\bfseries{43.9}} & 
					\textbf{\bfseries{24.2}} & 
					\textbf{\bfseries{45.2}} &
					\textbf{\bfseries{55.2}} 
					\\
					\midrule
					Mobile-Former-508M \cite{Mobile-former} & 
					T & 
					17.9M & 
					 - &
					38.0 & 
					58.3 & 
					40.3 & 
					22.9 & 
					41.2 & 
					49.7 
					\\
					\rowcolor{blue!10}
					\textbf{\bfseries{UniConvNet-P2(ours)}} & 
					\textbf{\bfseries{C}} & 
					\textbf{\bfseries{16.9M}} &
					\textbf{\bfseries{16.8G}} &
					\textbf{\bfseries{42.2}} & 
					\textbf{\bfseries{62.5}} & 
					\textbf{\bfseries{45.2}} & 
					\textbf{\bfseries{25.1}} & 
					\textbf{\bfseries{45.9}} &
					\textbf{\bfseries{56.2}} 
					\\
					\midrule
					PVTv1-Tiny \cite{Pvtv1} & 
					T & 
					23.0M & 
					 - &
					36.7 & 
					56.9 & 
					38.9 & 
					22.6 & 
					38.8 & 
					50.0 
					\\
					\rowcolor{yellow!10}
					ResNet-18 \cite{ResNet} & 
					C & 
					21.3M & 
					 - &
					31.8 & 
					49.6 & 
					33.6 & 
					16.3 & 
					34.3 & 
					43.2 
					\\
					
					\rowcolor{blue!10}
					\textbf{\bfseries{UniConvNet-N0(ours)}} & 
					\textbf{\bfseries{C}} & 
					\textbf{\bfseries{20.7M}} &
					\textbf{\bfseries{18.8G}} &
					\textbf{\bfseries{42.8}} & 
					\textbf{\bfseries{63.2}} & 
					\textbf{\bfseries{45.7}} & 
					\textbf{\bfseries{25.2}} & 
					\textbf{\bfseries{47.0}} &
					\textbf{\bfseries{56.8}} 
					\\
					\midrule
					PVTv2-B1 \cite{PVTv2} & 
					T & 
					23.8M & 
					 - &
					41.2 & 
					61.9 & 
					43.9 & 
					25.4 & 
					44.5 & 
					54.3 
					\\
					\rowcolor{blue!10}
					\textbf{\bfseries{UniConvNet-N1(ours)}} & 
					\textbf{\bfseries{C}} & 
					\textbf{\bfseries{23.9M}} &
					 \textbf{\bfseries{19.4G}} &
					\textbf{\bfseries{44.6}} & 
					\textbf{\bfseries{65.7}} & 
					\textbf{\bfseries{47.9}} & 
					\textbf{\bfseries{27.9}} & 
					\textbf{\bfseries{49.0}} &
					\textbf{\bfseries{59.2}} 
					\\
					\midrule
					PVTv1-Small \cite{Pvtv1} & 
					T & 
					44.1M & 
					 - &
					40.4 & 
					61.3 & 
					43.0 & 
					25.0 & 
					42.9 & 
					55.7 
					\\
					MPViT-T \cite{MPViT} & 
					T & 
					28.0M & 
					 - &
					41.8 & 
					62.7 & 
					44.6 & 
					27.2 & 
					45.1 & 
					54.2 
					\\
					Twins-PCPVT-S \cite{Twins-SVT} & 
					T & 
					34.4M & 
					 - &
					43.0 & 
					64.1 & 
					46.0 & 
					27.5 & 
					46.3 & 
					57.3 
					\\
					PVTv2-B2 \cite{PVTv2} & 
					T & 
					35.1M & 
					 - &
					44.6 & 
					65.6 & 
					47.6 & 
					27.4 & 
					48.8 & 
					58.6 
					\\
					Shunted-S \cite{ShuntedSA} & 
					T & 
					32.1M & 
					 - &
					45.4 & 
					65.9 & 
					49.2 & 
					28.7 & 
					49.3 & 
					60.0 
					\\
					\rowcolor{blue!10}
					\textbf{\bfseries{UniConvNet-N2(ours)}} & 
					\textbf{\bfseries{C}} & 
					\textbf{\bfseries{26.0M}} &
					 \textbf{\bfseries{20.7G}} &
					\textbf{\bfseries{45.5}} & 
					\textbf{\bfseries{66.4}} & 
					\textbf{\bfseries{48.9}} & 
					\textbf{\bfseries{28.4}} & 
					\textbf{\bfseries{50.2}} &
					\textbf{\bfseries{60.4}} 
					\\
					 \multicolumn{10}{c}{\textbf{   }}
					 \\
					 \multicolumn{10}{c}{\textbf{\bfseries{light SSDLite}}}
					 \\
					 \textbf{\bfseries{Backbone}} & 
					 \textbf{\bfseries{Type}}& 
					 \textbf{\bfseries{\#Params}}& 
					 \textbf{\bfseries{FLOPs}}&  
					 \textbf{$\boldsymbol{mAP}$} &
					 \textbf{$\boldsymbol{mAP_{50}}$} &
					 \textbf{$\boldsymbol{mAP_{75}}$} &
					 \textbf{$\boldsymbol{mAP_{S}}$} &
					 \textbf{$\boldsymbol{mAP_{M}}$} &
					 \textbf{$\boldsymbol{mAP_{L}}$}
					 \\
					 \midrule
					 \rowcolor{yellow!10}
					 MobileNetv3 \cite{MobileNetv3} & 
					 C & 
					 5.0M & 
					 0.6G & 
					 22.0 &
					 - & 
					 - & 
					 - & 
					 - & 
					 - 
					 \\
					 \rowcolor{yellow!10}
					 MobileNetv2 \cite{MobileNetv2} & 
					 C & 
					 4.3M & 
					 0.8G & 
					 22.1  &
					 - & 
					 - & 
					 - & 
					 - & 
					 - 
					 \\
					 \rowcolor{yellow!10}
					 MobileNetv1 \cite{MobileNetv1} & 
					 C & 
					 5.1M & 
					 1.3G & 
					 22.2 &
					 - & 
					 - & 
					 - & 
					 - & 
					 - 
					 \\
					 \midrule
					 MixNet \cite{MixNet} & 
					 T & 
					 4.5M & 
					 - & 
					 22.3 &
					 - & 
					 - & 
					 - & 
					 - & 
					 - 
					 \\
					 MNASNet \cite{MNASNet} & 
					 T & 
					 4.9M & 
					 0.8G & 
					 23.0 &
					 - & 
					 - & 
					 3.8 & 
					 21.7 & 
					 42.0 
					 \\
					 MobileViTv1-Small \cite{MobileViTv1} & 
					 T & 
					 5.7M & 
					 3.4G & 
					 27.7 &
					 - & 
					 - & 
					 - & 
					 - & 
					 - 
					 \\
					 EdgeNeXt-S \cite{EdgeNeXt} & 
					 T & 
					 6.2M & 
					 2.1G & 
					 27.9 &
					 - & 
					 - & 
					 - & 
					 - & 
					 - 
					 \\
					 \rowcolor{blue!10}
					 \textbf{\bfseries{UniConvNet-A(ours)}} & 
					 \textbf{\bfseries{C}} & 
					 \textbf{\bfseries{4.4M}} & 
					 \textbf{\bfseries{1.3G}} & 
					 \textbf{\bfseries{29.5}} & 
					 \textbf{\bfseries{46.7}} & 
					 \textbf{\bfseries{30.2}} & 
					 \textbf{\bfseries{5.3}} & 
					 \textbf{\bfseries{31.8}} & 
					 \textbf{\bfseries{53.6}} 
					 \\
					 \midrule
					 MobileViTv2-1.25 \cite{MobileViTv2} & 
					 T & 
					 8.2M & 
					 4.7G & 
					 27.8 &
					 - & 
					 - & 
					 - & 
					 - & 
					 - 
					 \\
					 EMO-5M \cite{EMO} & 
					 T & 
					 6.0M & 
					 1.8G & 
					 27.8 &
					 45.2 & 
					 28.2 & 
					 5.2 & 
					 30.5 & 
					 50.0 
					 \\
					 \rowcolor{blue!10}
					 \textbf{\bfseries{UniConvNet-P0(ours)}} & 
					 \textbf{\bfseries{C}} & 
					 \textbf{\bfseries{6.4M}} & 
					 \textbf{\bfseries{1.8G}} & 
					 \textbf{\bfseries{30.5}} & 
					 \textbf{\bfseries{48.1}} & 
					 \textbf{\bfseries{31.1}} & 
					 \textbf{\bfseries{6.2}} & 
					 \textbf{\bfseries{33.6}} & 
					 \textbf{\bfseries{55.3}} 
					 \\
					 \midrule
					 \rowcolor{blue!10}
					 \textbf{\bfseries{UniConvNet-P2(ours)}} & 
					 \textbf{\bfseries{C}} & 
					 \textbf{\bfseries{8.6M}} & 
					 \textbf{\bfseries{2.6G}} & 
					 \textbf{\bfseries{32.0}} & 
					 \textbf{\bfseries{49.8}} & 
					 \textbf{\bfseries{33.0}} & 
					 \textbf{\bfseries{7.6}} & 
					 \textbf{\bfseries{35.8}} & 
					 \textbf{\bfseries{56.0}} 
					 \\
					 \midrule
					 MobileViTv2-1.75 \cite{MobileViTv2} & 
					 T & 
					 14.9M & 
					 9.0G & 
					 29.5 &
					 - & 
					 - & 
					 - & 
					 - & 
					 - 
					 \\
					 \rowcolor{blue!10}
					 \textbf{\bfseries{UniConvNet-N0(ours)}} & 
					 \textbf{\bfseries{C}} & 
					 \textbf{\bfseries{12.2M}} & 
					 \textbf{\bfseries{3.8G}} & 
					 \textbf{\bfseries{33.7}} & 
					 \textbf{\bfseries{52.1}} & 
					 \textbf{\bfseries{34.8}} & 
					 \textbf{\bfseries{9.4}} & 
					 \textbf{\bfseries{37.6}} & 
					 \textbf{\bfseries{58.5}} 
					 \\
				\end{tabular}
		}
	\caption{\textbf{\bfseries{Object detection performance by heavy RetinaNet and light SSDLite on COCO val2017.}} “Type” refers to model type, where “C” and “T” denote pure CNN and the model using the Transformer for its architecture, respectively. The FLOPs are measured with $320 \times 320$ inputs.}
	\label{tab:table-RetinaNet-SSDLite}
\end{table}

\begin{table*}
	\belowrulesep=0pt
	\aboverulesep=0pt
	\centering
	\resizebox{0.75\linewidth}{!}
	{%
			\begin{tabular}{@{}lccccccccccccccc@{}}
					\multirow{2}{4em}{\textbf{\bfseries{Backbone}}} & 
					\multirow{2}{2em}{\textbf{\bfseries{Type}}} & 
					\multirow{2}{3em}{\textbf{\bfseries{\#Params}}} & 
					\multirow{2}{3em}{\textbf{\bfseries{FLOPs}}}&  
					\multicolumn{6}{c}{\textbf{Mask R-CNN $\boldsymbol{1\times}$ schedule}} &
					\multicolumn{6}{c}{\textbf{Mask R-CNN $\boldsymbol{3\times}$ + MS schedule}}
					\\
					&
					&
					&
					&
					\textbf{$\boldsymbol{AP^b}$} &
					\textbf{$\boldsymbol{AP^b_{50}}$} &
					\textbf{$\boldsymbol{AP^b_{75}}$} &
					\textbf{$\boldsymbol{AP^m}$} &
					\textbf{$\boldsymbol{AP^m_{50}}$} &
					\textbf{$\boldsymbol{AP^m_{75}}$} &
					\textbf{$\boldsymbol{AP^b}$} &
					\textbf{$\boldsymbol{AP^b_{50}}$} &
					\textbf{$\boldsymbol{AP^b_{75}}$} &
					\textbf{$\boldsymbol{AP^m}$} &
					\textbf{$\boldsymbol{AP^m_{50}}$} &
					\textbf{$\boldsymbol{AP^m_{75}}$} 
					\\
					\midrule
					\rowcolor{blue!10}
					\textbf{\bfseries{UniConvNet-N2}} &
					\textbf{\bfseries{C}} & 
					\textbf{\bfseries{34.7M}} & 
					\textbf{\bfseries{220G}} & 
					\textbf{\bfseries{46.6}} &
					\textbf{\bfseries{68.0}} &
					\textbf{\bfseries{51.3}} &
					\textbf{\bfseries{41.9}} &
					\textbf{\bfseries{65.1}} &
					\textbf{\bfseries{45.2}} &
					\textbf{\bfseries{48.4}} &
					\textbf{\bfseries{69.7}} &
					\textbf{\bfseries{53.2}} &
					\textbf{\bfseries{43.2}} &
					\textbf{\bfseries{66.7}} &
					\textbf{\bfseries{46.4}} 
					\\
					\midrule
					PVTv2-B2 \cite{PVTv2} &
					T &
					45.0M &
					309G &
					45.3 &
					67.1 &
					49.6 &
					41.2 &
					64.2 &
					44.4 &
					47.8 &
					69.7 &
					52.6 &
					43.1 &
					66.8 &
					46.7
					\\
					ViT-Adapter-S \cite{ViT-Adapter} &
					T &
					48.0M &
					403G &
					44.7 &
					65.8 &
					48.3 &
					39.9 &
					62.5 &
					42.8 &
					48.2 &
					69.7 &
					52.5 &
					42.8 &
					66.4 &
					45.9
					\\
					\rowcolor{yellow!10}
					MogaNet-S \cite{moganet} &
					C &
					45.0M &
					272G &
					46.7 &
					- &
					- &
					42.2 &
					- &
					- &
					- &
					- &
					- &
					- &
					- &
					-
					\\
					\rowcolor{blue!10}
					\textbf{\bfseries{UniConvNet-N3}} &
					\textbf{\bfseries{C}} & 
					\textbf{\bfseries{39.4M}} & 
					\textbf{\bfseries{239G}} & 
					\textbf{\bfseries{47.0}} & 
					\textbf{\bfseries{68.6}} & 
					\textbf{\bfseries{51.8}} & 
					\textbf{\bfseries{42.4}} & 
					\textbf{\bfseries{65.6}} & 
					\textbf{\bfseries{45.7}} &
					\textbf{\bfseries{49.4}} & 
					\textbf{\bfseries{70.7}} & 
					\textbf{\bfseries{54.4}} & 
					\textbf{\bfseries{44.2}} & 
					\textbf{\bfseries{67.9}} & 
					\textbf{\bfseries{47.5}}
					\\
					\midrule
					Swin-T \cite{Swin-transformer} &
					T &
					48.0M &
					267G &
					42.7 &
					65.2 &
					46.8 &
					39.3 &
					62.2 &
					42.2 &
					46.0 &
					68.1 &
					50.3 &
					41.6 &
					65.1 &
					44.9
					\\
					\rowcolor{yellow!10}
					ConvNeXt-T \cite{ConvNeXt} &
					C &
					48.0M &
					262G &
					44.2 &
					66.6 &
					48.3 &
					40.1 &
					63.3 &
					42.8 &
					46.2 &
					67.9 &
					50.8 &
					41.7 &
					65.0 &
					44.9
					\\
					\rowcolor{yellow!10}
					InternImage-T \cite{DCNV3} &
					C &
					49.0M &
					270G &
					47.2 &
					69.0 &
					52.1 &
					42.5 &
					66.1 &
					45.8 &
					49.1 &
					70.4 &
					54.1 &
					43.7 &
					67.3 &
					47.3
					\\
					\rowcolor{yellow!10}
					MogaNet-B \cite{moganet} &
					C &
					63.0M &
					373G &
					47.9 &
					- &
					- &
					43.2 &
					- &
					- &
					- &
					- &
					- &
					- &
					- &
					-
					\\
					\rowcolor{yellow!10}
					FlashInternImage-T \cite{DCNV4} &
					C &
					49.0M &
					- &
					48.0 &
					- &
					- &
					43.1 &
					- &
					- &
					49.5 &
					- &
					- &
					44.0 &
					- &
					-
					\\
					\rowcolor{blue!10}
					\textbf{\bfseries{UniConvNet-T}} &
					\textbf{\bfseries{C}} & 
					\textbf{\bfseries{50.0M}} & 
					\textbf{\bfseries{265G}} & 
					\textbf{\bfseries{48.2}} & 
					\textbf{\bfseries{69.8}} & 
					\textbf{\bfseries{52.9}} & 
					\textbf{\bfseries{43.3}} & 
					\textbf{\bfseries{66.6}} & 
					\textbf{\bfseries{45.7}} &
					\textbf{\bfseries{50.1}} & 
					\textbf{\bfseries{71.0}} & 
					\textbf{\bfseries{54.8}} & 
					\textbf{\bfseries{44.5}} & 
					\textbf{\bfseries{68.4}} & 
					\textbf{\bfseries{48.0}}
					\\
					\midrule
					Swin-S \cite{Swin-transformer} &
					T &
					69.0M &
					354G &
					44.8 &
					66.6 &
					48.9 &
					40.9 &
					63.4 &
					44.2 &
					48.2 &
					69.8 &
					52.8 &
					43.2 &
					67.0 &
					46.1
					\\
					\rowcolor{yellow!10}
					ConvNeXt-S \cite{ConvNeXt} &
					C &
					70.0M &
					348G &
					45.4 &
					67.9 &
					50.0 &
					41.8 &
					65.2 &
					45.1 &
					47.9 &
					70.0 &
					52.7 &
					42.9 &
					66.9 &
					46.2
					\\
					\rowcolor{yellow!10}
					InternImage-S \cite{DCNV3} &
					C &
					69.0M &
					340G &
					47.8 &
					69.8 &
					52.8 &
					43.3 &
					67.1 &
					46.7 &
					49.7 &
					71.1 &
					54.5 &
					44.5 &
					68.5 &
					47.8
					\\
					\rowcolor{yellow!10}
					FlashInternImage-S \cite{DCNV4} &
					C &
					69.0M &
					- &
					49.2 &
					- &
					- &
					44.0 &
					- &
					- &
					50.5 &
					- &
					- &
					44.9 &
					- &
					-
					\\
					\rowcolor{blue!10}
					\textbf{\bfseries{UniConvNet-S}} &
					\textbf{\bfseries{C}} & 
					\textbf{\bfseries{70.0M}} & 
					\textbf{\bfseries{336G}} & 
					\textbf{\bfseries{48.8}} & 
					\textbf{\bfseries{70.4}} & 
					\textbf{\bfseries{53.4}} & 
					\textbf{\bfseries{43.8}} & 
					\textbf{\bfseries{67.4}} & 
					\textbf{\bfseries{47.3}} &
					\textbf{\bfseries{50.8}} & 
					\textbf{\bfseries{71.6}} & 
					\textbf{\bfseries{55.6}} & 
					\textbf{\bfseries{45.2}} & 
					\textbf{\bfseries{69.3}} & 
					\textbf{\bfseries{48.9}}
					\\
					\midrule
					Swin-B \cite{Swin-transformer} &
					T &
					107.0M &
					496G &
					46.9 &
					- &
					- &
					42.3 &
					- &
					- &
					48.6 &
					70.0 &
					53.4 &
					43.3 &
					67.1 &
					46.7
					\\
					PVTv2-B5 \cite{PVTv2} &
					T &
					102.0M &
					557G &
					47.4 &
					68.6 &
					51.9 &
					42.5 &
					65.7 &
					46.0 &
					48.4 &
					69.2 &
					52.9 &
					42.9 &
					66.6 &
					46.2
					\\
					\rowcolor{yellow!10}
					ConvNeXt-B \cite{ConvNeXt} &
					C &
					108.0M &
					486G &
					47.0 &
					69.4 &
					51.7 &
					42.7 &
					66.3 &
					46.0 &
					48.5 &
					70.1 &
					53.3 &
					43.5 &
					67.1 &
					46.7
					\\
					\rowcolor{yellow!10}
					InternImage-B \cite{DCNV3} &
					C &
					115.0M &
					501G &
					48.8 &
					70.9 &
					54.0 &
					44.0 &
					67.8 &
					47.4 &
					50.3 &
					71.4 &
					55.3 &
					44.8 &
					68.7 &
					48.0
					\\
					\rowcolor{yellow!10}
					FlashInternImage-B \cite{DCNV4} &
					C &
					115.0M &
					- &
					50.1 &
					- &
					- &
					44.5 &
					- &
					- &
					50.6 &
					- &
					- &
					45.4 &
					- &
					-
					\\
					\rowcolor{blue!10}
					\textbf{\bfseries{UniConvNet-B}} &
					\textbf{\bfseries{C}} & 
					\textbf{\bfseries{118.0M}} & 
					\textbf{\bfseries{498G}} & 
					\textbf{\bfseries{50.0}} & 
					\textbf{\bfseries{71.7}} & 
					\textbf{\bfseries{55.3}} & 
					\textbf{\bfseries{45.0}} & 
					\textbf{\bfseries{69.0}} & 
					\textbf{\bfseries{48.5}} &
					\textbf{\bfseries{51.2}} & 
					\textbf{\bfseries{72.2}} & 
					\textbf{\bfseries{56.1}} & 
					\textbf{\bfseries{45.6}} & 
					\textbf{\bfseries{69.6}} & 
					\textbf{\bfseries{49.2}}
					\\
%
					\multicolumn{10}{c}{   }
					\\
					\textbf{\bfseries{Backbone}} & 
					\textbf{\bfseries{Type}} & 
					\textbf{\bfseries{\#Params}} & 
					\textbf{\bfseries{FLOPs}}&  
					\multicolumn{6}{c}{\textbf{Cascade Mask R-CNN $\boldsymbol{1\times}$ schedule}} &
					\multicolumn{6}{c}{\textbf{Cascade Mask R-CNN $\boldsymbol{3\times}$ + MS schedule}}
					\\
					\midrule
					Swin-L\dag \cite{Swin-transformer} &
					T &
					253.0M &
					1382G &
					51.8 &
					71.0 &
					56.2 &
					44.9 &
					68.4 &
					48.9 &
					53.9 &
					72.4 &
					58.8 &
					46.7 &
					70.1 &
					50.8
					\\
					\rowcolor{yellow!10}
					RepLKNet-31L\dag \cite{RepLKNet} &
					C &
					229.0M &
					1321G &
					- &
					- &
					- &
					- &
					- &
					- &
					53.9 &
					72.5 &
					58.6 &
					46.5 &
					70.0 &
					50.6
					\\
					\rowcolor{yellow!10}
					ConvNeXt-L\dag \cite{ConvNeXt} &
					C &
					255.0M &
					1354G &
					53.5 &
					72.8 &
					58.3 &
					46.4 &
					70.2 &
					50.2 &
					54.8 &
					73.8 &
					59.8 &
					47.6 &
					71.3 &
					51.7
					\\
					\rowcolor{yellow!10}
					ConvNeXt-XL\dag \cite{ConvNeXt} &
					C &
					407.0M &
					1898G &
					53.6 &
					72.9 &
					58.5 &
					46.5 &
					70.3 &
					50.5 &
					55.2 &
					74.2 &
					59.9 &
					47.7 &
					71.6 &
					52.2
					\\
					\rowcolor{yellow!10}
					HorNet-L\dag \cite{hornet} &
					C &
					259.0M &
					1358G &
					- &
					- &
					- &
					- &
					- &
					- &
					56.0 &
					- &
					- &
					48.6 &
					- &
					-
					\\
					
					\rowcolor{yellow!10}
					InternImage-L\dag \cite{DCNV3} &
					C &
					277.0M &
					1399G &
					54.9 &
					74.0 &
					59.8 &
					47.7 &
					71.4 &
					52.1 &
					56.1 &
					74.8 &
					60.7 &
					48.5 &
					72.4 &
					53.0
					\\
					\rowcolor{yellow!10}
					InternImage-XL\dag \cite{DCNV3} &
					C &
					387.0M &
					1782G &
					55.3 &
					74.4 &
					60.1 &
					48.1 &
					71.9 &
					52.4 &
					56.2 &
					75.0 &
					61.2 &
					48.8 &
					72.5 &
					53.4
					\\
					\rowcolor{yellow!10}
					FlashInternImage-L\dag \cite{DCNV4} &
					C &
					277.0M &
					- &
					55.6 &
					- &
					- &
					48.2 &
					- &
					- &
					56.7 &
					- &
					- &
					48.9 &
					- &
					-
					\\
					
					\rowcolor{blue!10}
					\textbf{\bfseries{UniConvNet-L\dag}} &
					\textbf{\bfseries{C}} & 
					\textbf{\bfseries{254.8M}} & 
					\textbf{\bfseries{1288G}} & 
					\textbf{\bfseries{55.7}} & 
					\textbf{\bfseries{74.4}} & 
					\textbf{\bfseries{60.4}} & 
					\textbf{\bfseries{48.3}} & 
					\textbf{\bfseries{71.9}} & 
					\textbf{\bfseries{52.9}} &
					\textbf{\bfseries{56.6}} & 
					\textbf{\bfseries{75.6}} & 
					\textbf{\bfseries{61.8}} & 
					\textbf{\bfseries{48.9}} & 
					\textbf{\bfseries{73.0}} & 
					\textbf{\bfseries{53.4}}
					\\
					 \bottomrule
				\end{tabular}
		}
	\caption{\textbf{\bfseries{Object detection and instance segmentation performance by Mask R-CNN and Cascade Mask R-CNN on COCO val2017.}} “Type” refers to model type, where “C” and “T” denote pure CNN and the model using the Transformer for its architecture, respectively. The FLOPs are measured with $1280\times800$ inputs.}
	\label{tab:table-Mask R-CNN}
\end{table*}

\begin{table}
	\belowrulesep=0pt
	\aboverulesep=0pt
	\centering
	\resizebox{0.7\linewidth}{!}
	{%
			\begin{tabular}{@{}lcccc@{}}
					\multicolumn{5}{c}{\textbf{\bfseries{DeepLabv3}}}
					\\
					\textbf{\bfseries{Model}} & 
					\textbf{\bfseries{Type}}& 
					\textbf{\bfseries{\#Params}}& 
					\textbf{\bfseries{FLOPs}}& 
					\textbf{\bfseries{mIoU}}
					\\
					\midrule
					MobileViTv2-0.5 \cite{MobileViTv2} & 
					T & 
					6.3M & 
					26.1G & 
					31.9 
					\\
					EMO-2M \cite{EMO} & 
					T & 
					6.9M & 
					3.5G & 
					35.3 
					\\
					\rowcolor{blue!10}
					\textbf{\bfseries{UniConvNet-A(ours)}} & 
					\textbf{\bfseries{C}} & 
					\textbf{\bfseries{7.9M}} & 
					\textbf{\bfseries{4.2G}} & 
					\textbf{\bfseries{38.2}} 
					\\
					\midrule
					MobileViTv2-0.75 \cite{MobileViTv2} & 
					T & 
					9.6M & 
					40.0G & 
					34.7 
					\\
					EMO-5M \cite{EMO} & 
					T & 
					10.3M & 
					5.8G & 
					37.8 
					\\
					\rowcolor{blue!10}
					\textbf{\bfseries{UniConvNet-P0(ours)}} & 
					\textbf{\bfseries{C}} & 
					\textbf{\bfseries{9.7M}} & 
					\textbf{\bfseries{5.3G}} & 
					\textbf{\bfseries{39.7}} 
					\\
					\midrule
					MobileViTv2-1.0 \cite{MobileViTv2} & 
					T & 
					13.4M & 
					56.4G & 
					37.0 
					\\
					\rowcolor{blue!10}
					\textbf{\bfseries{UniConvNet-P2(ours)}} & 
					\textbf{\bfseries{C}} & 
					\textbf{\bfseries{12.6M}} & 
					\textbf{\bfseries{7.8G}} & 
					\textbf{\bfseries{40.0}} 
					\\
					\midrule
					MobileNetv2 \cite{MobileNetv2} & 
					C & 
					18.7M & 
					75.4G & 
					34.1 
					\\
					\rowcolor{blue!10}
					\textbf{\bfseries{UniConvNet-N0(ours)}} & 
					\textbf{\bfseries{C}} & 
					\textbf{\bfseries{17.2M}} & 
					\textbf{\bfseries{11.0G}} & 
					\textbf{\bfseries{41.0}} 
					\\
					\midrule
					\rowcolor{blue!10}
					\textbf{\bfseries{UniConvNet-N1(ours)}} & 
					\textbf{\bfseries{C}} & 
					\textbf{\bfseries{20.3M}} & 
					\textbf{\bfseries{12.3G}} & 
					\textbf{\bfseries{42.1}} 
					\\
					\midrule
					MobileViTv2-2.0 \cite{MobileViTv2} & 
					T & 
					64.0M & 
					147.0G & 
					40.9 
					\\
					\rowcolor{yellow!10}
					ResNet-50 \cite{ResNet} & 
					C & 
					68.2M & 
					270.3G & 
					42.4 
					\\
					\rowcolor{blue!10}
					\textbf{\bfseries{UniConvNet-N2(ours)}} & 
					\textbf{\bfseries{C}} & 
					\textbf{\bfseries{22.5M}} & 
					\textbf{\bfseries{15.7G}} & 
					\textbf{\bfseries{42.9}} 
					\\
					\multicolumn{5}{c}{   }
					\\
					\multicolumn{5}{c}{\textbf{\bfseries{PSPNet}}}
					\\
					\textbf{\bfseries{Model}} & 
					\textbf{\bfseries{Type}}& 
					\textbf{\bfseries{\#Params}}& 
					\textbf{\bfseries{FLOPs}}& 
					\textbf{\bfseries{mIoU}}
					\\
					\midrule
					EMO-2M \cite{EMO} & 
					T & 
					5.5M & 
					3.1G & 
					34.5 
					\\
					MobileViTv2-0.75 \cite{MobileViTv2} & 
					T & 
					6.2M & 
					26.6G & 
					35.2 
					\\
					\rowcolor{blue!10}
					\textbf{\bfseries{UniConvNet-A(ours)}} & 
					\textbf{\bfseries{C}} & 
					\textbf{\bfseries{6.5M}} & 
					\textbf{\bfseries{3.8G}} & 
					\textbf{\bfseries{37.9}} 
					\\
					\midrule
					MobileViTv2-1.0 \cite{MobileViTv2} & 
					T & 
					9.4M & 
					40.3G & 
					36.5 
					\\
					EMO-5M \cite{EMO} & 
					T & 
					8.5M & 
					5.3G & 
					38.2 
					\\
					\rowcolor{blue!10}
					\textbf{\bfseries{UniConvNet-P0(ours)}} & 
					\textbf{\bfseries{C}} & 
					\textbf{\bfseries{8.1M}} & 
					\textbf{\bfseries{4.8G}} & 
					\textbf{\bfseries{39.2}} 
					\\
					\midrule
					MobileNetv2 \cite{MobileNetv2} & 
					T & 
					13.7M & 
					53.1G & 
					29.7 
					\\
					\rowcolor{blue!10}
					\textbf{\bfseries{UniConvNet-P2(ours)}} & 
					\textbf{\bfseries{C}} & 
					\textbf{\bfseries{10.9M}} & 
					\textbf{\bfseries{7.3G}} & 
					\textbf{\bfseries{39.6}} 
					\\
					\midrule
					\rowcolor{blue!10}
					\textbf{\bfseries{UniConvNet-N0(ours)}} & 
					\textbf{\bfseries{C}} & 
					\textbf{\bfseries{15.0M}} & 
					\textbf{\bfseries{10.4G}} & 
					\textbf{\bfseries{40.1}} 
					\\
					\midrule
					MobileViTv2-1.75 \cite{MobileViTv2} & 
					T & 
					22.5M & 
					95.9G & 
					39.8 
					\\
					\rowcolor{blue!10}
					\textbf{\bfseries{UniConvNet-N1(ours)}} & 
					\textbf{\bfseries{C}} & 
					\textbf{\bfseries{18.2M}} & 
					\textbf{\bfseries{11.6G}} & 
					\textbf{\bfseries{42.1}} 
					\\
					\midrule
					ResNet-50 \cite{ResNet} & 
					T & 
					49.1M & 
					179.1G & 
					41.1 
					\\
					\rowcolor{blue!10}
					\textbf{\bfseries{UniConvNet-N2(ours)}} & 
					\textbf{\bfseries{C}} & 
					\textbf{\bfseries{20.3M}} & 
					\textbf{\bfseries{15.1G}} & 
					\textbf{\bfseries{42.5}} 
					\\
				\end{tabular}
		}
	\caption{\textbf{\bfseries{Semantic segmentation performance by DeepLabv3 and PSPNet on ADE20K dataset.}} “Type” refers to model type, where “C” and “T” denote pure CNN and the model using the Transformer for its architecture, respectively. The FLOPs are measured with $512 \times 512$ inputs.}
	\label{tab:Semantic segmentation DeepLabv3}
\end{table}

\begin{table}
\belowrulesep=0pt
\aboverulesep=0pt
	\centering
	\resizebox{0.85\linewidth}{!}
	{%
			\begin{tabular}{@{}lcccccc@{}}
					\textbf{\bfseries{Backbone}} & 
					\textbf{\bfseries{Type}} & 
					\textbf{\bfseries{Crop Size}} & 
					\textbf{\bfseries{\#Params}} & 
					\textbf{\bfseries{FLOPs}} & 
					\textbf{\bfseries{mIoU(SS)}} & 
					\textbf{\bfseries{mIoU(MS)}}
					\\
					\midrule
					Swin-T \cite{Swin-transformer} &
					T &
					512\textsuperscript{2} & 
					60.0 &
					939 &
					44.5 &
					45.8 
					\\
					\rowcolor{yellow!10}
					ConvNeXt-T \cite{ConvNeXt} &
					C &
					512\textsuperscript{2} & 
					60.0&
					939 &
					46.0 &
					46.7 
					\\
					\rowcolor{blue!10}
					\textbf{\bfseries{UniConvNet-N2}} &
					\textbf{\bfseries{C}} & 
					\textbf{\bfseries{512\textsuperscript{2}}} & 
					\textbf{\bfseries{43.9}} & 
					\textbf{\bfseries{893}} &
					\textbf{\bfseries{47.9}} &
					\textbf{\bfseries{48.9}}
					\\
					\midrule
					\rowcolor{yellow!10}
					SLaK-T \cite{SLaK} &
					C &
					512\textsuperscript{2} & 
					65.0&
					936 &
					47.6 &
					- 
					\\
					\rowcolor{yellow!10}
					InternImage-T \cite{DCNV3} &
					C &
					512\textsuperscript{2} & 
					59.0&
					944 &
					47.9 &
					48.1 
					\\
					\rowcolor{blue!10}
					\textbf{\bfseries{UniConvNet-N3}} &
					\textbf{\bfseries{C}} & 
					\textbf{\bfseries{512\textsuperscript{2}}} & 
					\textbf{\bfseries{48.6}} & 
					\textbf{\bfseries{912}} &
					\textbf{\bfseries{49.0}} &
					\textbf{\bfseries{50.0}}
					\\
					\midrule
					Shunted-S \cite{ShuntedSA} &
					T &
					512\textsuperscript{2} & 
					52.0&
					940 &
					48.9 &
					49.9 
					\\
					\rowcolor{yellow!10}
					PeLK-T \cite{pelk} &
					C &
					512\textsuperscript{2} & 
					62.0&
					970 &
					48.1 &
					- 
					\\
					\rowcolor{yellow!10}
					UniRepLKNet-T \cite{UniRepLKNet} &
					C &
					512\textsuperscript{2} & 
					61.0&
					946 &
					48.6 &
					49.1 
					\\
					\rowcolor{yellow!10}
					FlashInternImage-T \cite{DCNV4} &
					C &
					512\textsuperscript{2} & 
					59.0&
					944 &
					49.3 &
					50.3 
					\\
					\rowcolor{yellow!10}
					MogaNet-B \cite{moganet} &
					C &
					512\textsuperscript{2} & 
					74.0&
					1050 &
					50.1 &
					- 
					\\
					\rowcolor{blue!10}
					\textbf{\bfseries{UniConvNet-T}} &
					\textbf{\bfseries{C}} & 
					\textbf{\bfseries{512\textsuperscript{2}}} & 
					\textbf{\bfseries{59.2}} & 
					\textbf{\bfseries{939}} &
					\textbf{\bfseries{50.3}} &
					\textbf{\bfseries{51.2}}
					\\
					\midrule
					Swin-S \cite{Swin-transformer} &
					T &
					512\textsuperscript{2} & 
					81.0 &
					1038 &
					47.6 &
					49.5 
					\\
					\rowcolor{yellow!10}
					ConvNeXt-S \cite{ConvNeXt} &
					C &
					512\textsuperscript{2} & 
					82.0&
					7027 &
					48.7 &
					49.6
					\\
					\rowcolor{yellow!10}
					SLaK-S \cite{SLaK} &
					C &
					512\textsuperscript{2} & 
					91.0&
					1028 &
					49.4 &
					- 
					\\
					\rowcolor{yellow!10}
					PeLK-S \cite{pelk} &
					C &
					512\textsuperscript{2} & 
					84.0&
					1077 &
					49.7 &
					- 
					\\
					\rowcolor{yellow!10}
					InternImage-S \cite{DCNV3} &
					C &
					512\textsuperscript{2} & 
					80.0&
					1017 &
					50.1 &
					50.9 
					\\
					\rowcolor{yellow!10}
					UniRepLKNet-S \cite{UniRepLKNet} &
					C &
					512\textsuperscript{2} & 
					86.0&
					1036 &
					50.5 &
					51.0 
					\\
					\rowcolor{yellow!10}
					FlashInternImage-S \cite{DCNV3} &
					C &
					512\textsuperscript{2} & 
					80.0&
					- &
					50.6 &
					51.6 
					\\
					\rowcolor{yellow!10}
					MogaNet-L \cite{moganet} &
					C &
					512\textsuperscript{2} & 
					113.0&
					1176 &
					50.9 &
					- 
					\\
					\rowcolor{blue!10}
					\textbf{\bfseries{UniConvNet-S}} &
					\textbf{\bfseries{C}} & 
					\textbf{\bfseries{512\textsuperscript{2}}} & 
					\textbf{\bfseries{78.9}} & 
					\textbf{\bfseries{1015}} &
					\textbf{\bfseries{52.2}} &
					\textbf{\bfseries{52.8}}
					\\
					\midrule
					Swin-B \cite{Swin-transformer} &
					T &
					512\textsuperscript{2} & 
					121.0 &
					1188 &
					48.1 &
					49.7 
					\\
					\rowcolor{yellow!10}
					ConvNeXt-B \cite{ConvNeXt} &
					C &
					512\textsuperscript{2} & 
					122.0&
					1170 &
					49.1 &
					49.9 
					\\
					\rowcolor{yellow!10}
					RepLKNet-31B \cite{RepLKNet} &
					C &
					512\textsuperscript{2} & 
					112.0&
					1170 &
					49.9 &
					50.6 
					\\
					\rowcolor{yellow!10}
					SLaK-B \cite{SLaK} &
					C &
					512\textsuperscript{2} & 
					135.0&
					1172 &
					50.2 &
					- 
					\\
					\rowcolor{yellow!10}
					PeLK-B-101 \cite{pelk} &
					C &
					512\textsuperscript{2} & 
					126.0&
					1339 &
					50.6 &
					- 
					\\
					\rowcolor{yellow!10}
					InternImage-B \cite{DCNV3} &
					C &
					512\textsuperscript{2} & 
					128.0&
					1185 &
					50.8 &
					51.3 
					\\
					\rowcolor{yellow!10}
					FlashInternImage-B \cite{DCNV3} &
					C &
					512\textsuperscript{2} & 
					128.0&
					- &
					52.0 &
					52.6 
					\\
					\rowcolor{blue!10}
					\textbf{\bfseries{UniConvNet-B}} &
					\textbf{\bfseries{C}} & 
					\textbf{\bfseries{512\textsuperscript{2}}} & 
					\textbf{\bfseries{126.5}} & 
					\textbf{\bfseries{1179}} &
					\textbf{\bfseries{52.3}} &
					\textbf{\bfseries{52.9}}
					\\
					\midrule
					Swin-L\dag \cite{Swin-transformer} &
					T &
					640\textsuperscript{2} & 
					234 &
					2468 &
					52.1 &
					53.5 
					\\
					\rowcolor{yellow!10}
					RepLKNet-31L\dag \cite{RepLKNet} &
					C &
					640\textsuperscript{2} & 
					207&
					2404 &
					52.4 &
					52.7 
					\\
					\rowcolor{yellow!10}
					\rowcolor{yellow!10}
					ConvNeXt-L\dag \cite{ConvNeXt} &
					C &
					640\textsuperscript{2} & 
					235 &
					2458 &
					53.2 &
					53.7 
					\\
					\rowcolor{yellow!10}
					ConvNeXt-XL\dag \cite{ConvNeXt} &
					C &
					640\textsuperscript{2} & 
					391 &
					3335 &
					53.6 &
					54.0 
					\\
					\rowcolor{yellow!10}
					InternImage-L\dag \cite{DCNV3} &
					C &
					640\textsuperscript{2} & 
					256 &
					2526 &
					53.9 &
					54.1 
					\\
					\rowcolor{yellow!10}
					InternImage-XL\dag \cite{DCNV3} &
					C &
					640\textsuperscript{2} & 
					368 &
					3142 &
					55.0 &
					55.3 
					\\
					\rowcolor{blue!10}
					\textbf{\bfseries{UniConvNet-L\dag}} &
					\textbf{\bfseries{C}} & 
					\textbf{\bfseries{640\textsuperscript{2}}} & 
					\textbf{\bfseries{234}} & 
					\textbf{\bfseries{2310}} &
					\textbf{\bfseries{55.1}} &
					\textbf{\bfseries{55.4}}
					\\
				\end{tabular}
		}
	\caption{\textbf{\bfseries{Semantic segmentation performance by UperNet on ADE20K validation set.}} “Type” refers to model type, where “C” and “T” denote pure CNN and the model using the Transformer for its architecture, respectively. “SS” and “MS” denote single-scale and multi-scale testing, respectively. The FLOPs are measured with $512 \times 2048$ or $640 \times 2560$ inputs.}
\label{tab:table-UperNet}
\end{table}

\subsection{Downstream Tasks}

\paragraph{Object Detection and Instance Segmentation on COCO}

We fine-tune the heavy RetinaNet \cite{RetinaNet} and light SSDLite \cite{MobileNetv3} using our ImageNet-1K pre-trained UniConvNet on the MS-COCO 2017 with a $1\times$ schedule training recipe. 
We further fine-tune scaled-up UniConvNet using the representative object detection framework Mask R-CNN \cite{Mask-RCNN} and Cascade Mask R-CNN \cite{Cascade} on the MS-COCO 2017 datasets, employing either a $1\times$ (12epoch) training schedule or a $3\times$ (36 epoch) training schedule. Detailed fine-tuning settings are provided in \cref{Object Detection and Instance Segmentation Fine-tune}. 

\paragraph{Semantic Segmentation on ADE20K}

We fine-tune DeepLabv3 \cite{DeepLabv3} and PSPNet \cite{PSPNet} using our ImageNet-1K pre-trained UniConvNet on the ADE20K dataset with a training recipe of 160k iterations. 
We also fine-tune the scaled-up UniConvNet using the representative semantic segmentation framework UperNet on the ADE20K dataset for 160k iterations. 
Detailed fine-tuning settings are provided in \cref{Semantic Segmentation Fine-tune}.

\paragraph{Overall Results}
As shown in \cref{tab:table-RetinaNet-SSDLite}, \cref{tab:table-Mask R-CNN}, \cref{tab:Semantic segmentation DeepLabv3} and \cref{tab:table-UperNet}, our proposed UniConvNet significantly enhances performance compared to state-of-the-art models, offering lighter parameters and reduced FLOPs. This demonstrates the effectiveness and efficiency of the proposed Three-layer RFA and its improved capability for downstream tasks.

%

		%
\section{Analysis}
\label{sec:Analysis}

\subsection{Three-layer RFA properly expand ERF while maintaining AGD}
\label{sec:Ablation Study}


\begin{table}
	\belowrulesep=0pt
	\aboverulesep=0pt
	\centering
	\resizebox{0.75\linewidth}{!}
	{%
		\begin{tabular}{@{}lccccc@{}}
			\textbf{\bfseries{{Model}}} & 
			\textbf{\bfseries{{N}}} & 
			\textbf{\bfseries{Kernel Size}} & 
			\textbf{\bfseries{\#Params}} &  
			\textbf{\bfseries{FLOPs}}& 
			\textbf{\bfseries{Acc(\%)}}
			\\
			\midrule
			& 
			3 & 
			5, 7, 9 &
			3.5M &
			0.564G &
			76.6 
			\\
			\rowcolor{blue!10}
			UniConvNet-A &
			3 & 
			7, 9, 11 &
			3.4M &
			0.589G &
			77.0 
			\\
			&
			3 & 
			9, 11, 13 &
			3.5M &
			0.579G &
			76.9
			\\
			\midrule
			& 
			3 & 
			5, 7, 9 &
			5.1M &
			0.810G &
			78.8 
			\\
			\rowcolor{blue!10}
			UniConvNet-P0 &
			3 & 
			7, 9, 11 &
			5.2M &
			0.832G &
			79.1 
			\\
			&
			3 & 
			9, 11, 13 &
			5.3M &
			0.868G &
			79.3
			\\
			&
			3 & 
			11, 13, 15 &
			5.1M &
			0.845G &
			78.8
			\\
			\midrule
			& 
			3 & 
			5, 7, 9 &
			30.0M &
			5.0G &
			84.1 
			\\
			\rowcolor{blue!10}
			UniConvNet-T &
			3 & 
			7, 9, 11 &
			30.3M &
			5.1G &
			84.2
			\\
			&
			3 & 
			9, 11, 13 &
			29.6M &
			5.0G &
			84.1
			\\
			\midrule
			\rowcolor{blue!10}
			UniConvNet-N0 &
			3 & 
			7, 9, 11 &
			10.2M &
			1.65G &
			81.6 
			\\
			&
			4 & 
			5, 7, 9, 11 &
			9.8M &
			1.64G &
			81.5 
			\\
			&
			4 & 
			7, 9, 11, 13 &
			10.0M &
			1.70G &
			81.3 
			\\
			\midrule
			&
			3 & 
			27, 29, 31 &
			14.9M &
			2.87G &
			81.8 
			\\
			\rowcolor{blue!10}
			UniConvNet-N2 &
			3 & 
			7, 9, 11 &
			15.0M &
			2.47G &
			82.7 
			\\
			\midrule
			\rowcolor{blue!10}
			&
			3 & 
			7, 11, 31 &
			3.56M &
			0.687G &
			76.0
			\\
			&
			3 & 
			7, 9, 29 &
			3.45M &
			0.647G &
			75.9 
			\\
			UniConvNet-A &
						3 & 
			7, 9, 11 &
			3.4M &
			0.589G &
			77.0 
			\\
			\midrule
			&
						3 & 
			7, 11, 31 &
			6.2M &
			1.162G &
			79.4
			\\
			\rowcolor{blue!10}
			UniConvNet-P1 &
			3 & 
			7, 9, 11 &
			6.1M &
			0.895G &
			79.6 
			\\
		\end{tabular}
	}
	\caption{\textbf{\bfseries{Ablation studies on layer number and progressive kernel size.}} “Acc” is the TOP-1 accuracy. “N” represents the number in the pyramid order.}
	\label{tab:Ablation on pyramid order and Kernel Size}
\end{table}

\paragraph{Ablation Studies on Layer Number and Kernel Size}\label{Ablation on pyramid order and Kernel Size}

\Cref{Three-layer RFA for UniConvNet} 
presents the constraints on layer $N$ and progressive kernel size.
Based on UniConvNet-T, we adjust layer $N$ and progressive kernel size for ablation studies.
The results indicate that the proposed principles for these two hyperparameters are effective and accurate.
As illustrated in \cref{tab:Ablation on pyramid order and Kernel Size}, for layer $3$, progressive kernel sizes smaller than $(5, 7, 9)$ may be insufficient for expanding ERF to a level of existing large-kernel ConvNets and result in inferior performance.
Progressive kernel sizes larger than $(9, 11, 13)$ achieve better TOP-1 accuracy than UniConvNet-P0, with slightly higher parameters and FLOPs. In contrast, UniConvNet-A and UniConvNet-T perform better with kernel sizes of $(7, 9, 11)$.
Progressive kernel sizes of $(9, 11, 13)$ are inefficient for constructing deep models with equivalent parameters, which is crucial for model perception.
Therefore, we choose a kernel size of $(7, 9, 11)$ for efficiency.
For layer $4$, progressive kernels result in a theoretical receptive field much larger than the image size of $14\times{14}$ in stage $3$, which is wasteful for an image resolution of $224\times{224}$. It contradicts the original goal of alleviating parameters and FLOPs burden in contemporary large-kernel ConvNets.

\paragraph{Investigation on different AGD of ERF}\label{Investigation on different AGD of ERF}
We also examine the perceptual capabilities of large kernel sizes such as $(27, 29, 31)$ as used in RepLKNet \cite{RepLKNet}.
The performance results show that using large kernels is neither efficient nor effective for constructing long-range ERF following AGD, disrupting AGD on small-scale pixels.
We further use relatively small kernels in the first two layers, then use an extremely large kernel in the latter layer, such as $(7, 9, 29)$ and $(7, 11, 31)$.
This establishes a small-scale AGD in a smaller area, compared to the kernel sizes of $(7, 9, 11)$, and expands the ERF by the latter extremely large kernel.
The inferior performance demonstrates that a large ERF with continuous AGD, generated by three-layer RFA, from center to edge is vital and proper.
We analyze the ERF of several models and demonstrate that a proper AGD of small-scale pixels is more important than expanding the ERF. Please refer to \cref{Effective Receptive Field and Multi-scale perception ability} for detailed analysis.


\subsection{Three-layer RFA is Efficient and Effective}
\label{Efficiency and Effectiveness of Three-layer RFA}

\begin{table}
	\belowrulesep=0pt
	\aboverulesep=0pt
	\centering
	\resizebox{\linewidth}{!}
	{
	\begin{tabular}{@{}l|c|c|c|c|c@{}}
\multirow{2}{4em}{\textbf{\bfseries{UniConvNet}}} & 
\multirow{2}{3em}{\textbf{\bfseries{Overall}}} & 
\multirow{2}{3em}{\textbf{\bfseries{RFA$^{3-l}$}}}& 
\multirow{2}{3.5em}{\textbf{\bfseries{Modified DCNV3}}}&  
\multirow{2}{3em}{\textbf{\bfseries{Feed-Forward}}}& 
\multirow{2}{3em}{\textbf{\bfseries{Class Head}}}
\\
&
&
&
&
&

\\
\midrule
UniConvNet-A & 
3.4/0.589 &
\textbf{\bfseries{0.747/0.157}} &
\textbf{\bfseries{0.91/0.161}} &
1.331/0.234 &
0.409/0.037
\\
UniConvNet-P0 & 
5.17/0.832 &
\textbf{\bfseries{1.06/0.205}} &
\textbf{\bfseries{1.41/0.229}} &
2.06/0.334 &
0.64/0.066
\\
UniConvNet-P1 & 
6.1/0.895 &
\textbf{\bfseries{1.26/0.22}} &
\textbf{\bfseries{1.70/0.247}} &
2.495/0.362 &
0.635/0.064
\\
UniConvNet-P2 & 
7.57/1.25 &
\textbf{\bfseries{1.625/0.315}} &
\textbf{\bfseries{2.16/0.354}} &
3.153/0.516 &
0.648/0.065
\\
UniConvNet-N0 & 
10.23/1.65 &
\textbf{\bfseries{2.18/0.405}} &
\textbf{\bfseries{2.16/0.354}} &
4.62/0.851 &
1.27/0.04
\\
UniConvNet-N1 & 
13.06/1.88 &
\textbf{\bfseries{2.85/0.458}} &
\textbf{\bfseries{2.85/0.409}} &
6.095/0.867 &
1.275/0.134
\\
UniConvNet-N2 & 
15.0/2.47 &
\textbf{\bfseries{3.34/0.622}} &
\textbf{\bfseries{3.31/0.546}} &
7.08/1.161 &
1.27/0.139
\\
UniConvNet-N3 & 
19.7/3.37 &
\textbf{\bfseries{4.54/0.849}} &
\textbf{\bfseries{4.45/0.762}} &
9.48/1.619 &
1.25/0.142
\\
UniConvNet-T & 
30.3/5.1 &
\textbf{\bfseries{5.41/1.06}} &
\textbf{\bfseries{5.74/1.0}} &
14.61/2.46 &
4.54/0.58
\\
UniConvNet-S & 
50/8.48 &
\textbf{\bfseries{4.54/0.849}} &
\textbf{\bfseries{4.45/0.762}} &
9.48/1.619 &
1.25/0.142
\\
UniConvNet-B & 
97.6/15.9 &
\textbf{\bfseries{10.89/1.97}} &
\textbf{\bfseries{11.44/1.95}} &
24.61/4.18 &
3.06/0.38
\\
UniConvNet-L & 
201.8/100.1 &
\textbf{\bfseries{22.01/21.79}} &
\textbf{\bfseries{25.77/23.48}} &
80.24/50.56 &
11.11/4.28
\\
UniConvNet-XL & 
226.7/115.2 &
\textbf{\bfseries{24.7/25.07}} &
\textbf{\bfseries{28.95/27.02}} &
90.14/58.19 &
12.48/4.93
\\
\end{tabular}
}
\caption{\textbf{\bfseries{\#Parameters and FLOPs distributions of UniConvNet variants.}} \#Parameters(M)/FLOPs(G) are \#parameters and FLOPs for each block, respectively (\eg, 3.4/0.589 are the overall \#parameters and FLOPs for UniConvNet-A, respectively).}
\label{tab:param-FLOPs-distribution}
\end{table}

\paragraph{Efficiency}\label{Distributions-of-param&FLOPs}

We present the parameters and FLOPs of UniConvNet variants in \cref{tab:param-FLOPs-distribution}.
UniConvNet comprises Three-layer RFA, modified DCNV3, a feed-forward layer, and a class head.
Generally, compared to small-kernel modified DCNV3, Three-layer RFA has fewer or comparable parameters and FLOPs.
This suggests that our proposed Three-layer RFA can establish long-range dependencies while reducing FLOP costs and enhancing parameter efficiency.

\begin{table}
\belowrulesep=0pt
\aboverulesep=0pt
\centering
\resizebox{0.9\linewidth}{!}
{%
\begin{tabular}{@{}l|c|cc|c@{}}
\textbf{\bfseries{{Model}}} & 
\textbf{\bfseries{\#Params}} & 
\textbf{\bfseries{Large Kernel}} & 
\textbf{\bfseries{Small Kernel}} & 
\textbf{\bfseries{Acc(\%)}}
\\
\midrule
\rowcolor{blue!10}
UniConvNet-P0 & 
5.2M &
Three-layer RFA &
Modified DCNV3 &
79.1 
\\
& 
5.2M &
Three-layer RFA &
\usym{2613}&
78.4 
\\
\rowcolor{blue!5}
FlashInternImage \cite{DCNV4} & 
5.3M &
\usym{2613}&
DCNV4 &
78.5 
\\
& 
5.1M &
Three-layer RFA &
DW $3\times3$ &
78.9 
\\
\rowcolor{blue!5}
ConvNeXt \cite{ConvNeXt} & 
5.2M &
DW $7\times7$ &
DW $3\times3$ &
77.0 
\\
& 
5.3M &
\usym{2613}&
DW $3\times3$ &
77.0 
\\
\midrule
\rowcolor{blue!10}
UniConvNet-N2 & 
15.0M &
Three-layer RFA &
Modified DCNV3 &
82.7 
\\
\rowcolor{blue!5}
FlashInternImage \cite{DCNV4} & 
15.3M &
\usym{2613}&
DCNV4 &
82.2
\\
& 
14.9M &
Three-layer RFA &
DW $3\times3$ &
81.9 
\\
\rowcolor{blue!5}
ConvNeXt \cite{ConvNeXt} & 
15.1M &
DW $7\times7$ &
DW $3\times3$ &
81.0 
\\
\midrule
\rowcolor{blue!10}
UniConvNet-T & 
30.3M &
Three-layer RFA &
Modified DCNV3 &
84.2
\\
\rowcolor{blue!5}
FlashInternImage-T \cite{DCNV4} & 
30.0M &
\usym{2613}&
DCNV4 &
83.6
\\
& 
30.7M &
Three-layer RFA &
DW $3\times3$ &
83.7 
\\
\rowcolor{blue!5}
ConvNeXt-T \cite{ConvNeXt} & 
29.0M &
DW $7\times7$ &
DW $3\times3$ &
82.1 
\\
\end{tabular}
}
\caption{\textbf{\bfseries{Ablation comparisons of different large-kernel and small-kernel convolutions.}} “Acc” is the TOP-1 accuracy.}
\label{tab:Effectiveness of Multi-head Recursive Convolution}
\end{table}

\paragraph{Effectiveness}\label{Effectiveness-of-Multi-head Recursive Convolution}

We perform an ablation study comparing various combinations of large-kernel and small-kernel convolutions to evaluate the effectiveness of different modules.
As illustrated in \cref{tab:Effectiveness of Multi-head Recursive Convolution}, models using only Three-layer RFA or Modified DCNV3 achieve similar top-1 accuracies of 78.4 and 78.5, respectively. This indicates that the proposed Three-layer RFA retains comparable feature perception capabilities without relying on the basic small-scale information typically used in conventional ConvNets.
Additionally, we replace Modified DCNV3 in UniConvNet-P0 with a depth-wise $3\times3$ convolution.
The top-1 accuracy is 78.9, 0.2 lower than UniConvNet-P0, indicating that UniConvNet-P0 is slightly improved by Modified DCNV3 compared to when it uses simple depth-wise $3\times3$ convolution.
In contrast, replacing Three-layer RFA with another large-kernel convolution (depth-wise $7\times7$ convolution) results in a drop in top-1 accuracy to 77.0, highlighting the effectiveness of our proposed Three-layer RFA.
Furthermore, models using only depth-wise $3\times3$ convolution achieve a top-1 accuracy of 77.0, indicating that depth-wise $7\times7$ convolution does not enhance the model's perception capabilities.
This further validates the effectiveness of the proposed Three-layer RFA.
The ablation studies on models of $15M$ and $30M$ parameters manifest a consistent performance improvement ability of the proposed Three-layer RFA.

\begin{table}
	\belowrulesep=0pt
	\aboverulesep=0pt
	\centering
	\resizebox{0.8\linewidth}{!}
	{%
		\begin{tabular}{@{}lcccc@{}}
			\textbf{\bfseries{{Model}}} & 
			\textbf{\bfseries{\#Params}} & 
			\textbf{\bfseries{Scale}} &  
			\textbf{\bfseries{Acc(\%)}} &  
			\textbf{\bfseries{Throughput}} 
			\\
			\midrule
			Swin-T \cite{Swin-transformer} & 
			29.0M &
			224\textsuperscript{2} &
			81.3&
			1989/3619 
			\\
			ConvNeXt-T \cite{ConvNeXt} & 
			29.0M &
			224\textsuperscript{2} &
			82.1&
			2485/4305 
			\\
			\rowcolor{blue!6}
			InternImage-T \cite{DCNV3} & 
			30.0M &
			224\textsuperscript{2} &
			83.5&
			1409/1746 
			\\
			\rowcolor{blue!12}
			UniConvNet-T & 
			30.3M &
			224\textsuperscript{2} &
			84.2&
			1480/1825 
			\\
			\midrule
			ConvNeXt-XL \cite{ConvNeXt} & 
			350.0M &
			224\textsuperscript{2} &
			87.7&
			170/299 
			\\
			\rowcolor{blue!6}
			InternImage-XL \cite{DCNV3} & 
			335.0M &
			224\textsuperscript{2} &
			88.0&
			125/174 
			\\
			\rowcolor{blue!6}
			InternImage-L \cite{DCNV3} & 
			223.0M &
			224\textsuperscript{2} &
			87.7&
			158/214 
			\\
			\rowcolor{blue!12}
			UniConvNet-XL & 
			226.7M & 
			224\textsuperscript{2} &
			88.4&
			168/228 
			\\
		\end{tabular}
	}
	\caption{\textbf{\bfseries{Image classification throughput on ImageNet-1K.}} “Acc(\%)” is the TOP-1 Accuracy. The overall throughput of each model, measured as the number of images processed per second, is reported in FP32/FP16 data formats.}
	\label{tab:Image classification throughput on ImageNet-1K-main}
\end{table}

\subsection{Throughput analysis}\label{Throughput analysis-main}

We benchmark throughput on several classic and relevant models. 
As shown in \cref{tab:Image classification throughput on ImageNet-1K-main}, UniConvNets improve the throughput compared to InternImage \cite{DCNV3} with similar $3\times{3}$ convolution.
Please refer to \cref{Throughput analysis} for detailed configurations.

\section{Related Work}




Convolutional neural networks (ConvNets) \cite{mip, ResNet, cardinality, depth, width, Res2net, ResNeSt, repvgg, DCNV3, DCNV4, MobileNetv1, MobileNetv2, MobileNetv3, StarNet} have long been the standard architecture for vision recognition due to their intrinsic inductive bias.
CNNs utilize a stack of small kernels to establish local dependencies, which limits their ability to perceive.
However, their dominance is being challenged by the emergence of attention mechanisms in computer vision.
In recent years, attention-based models \cite{ViT, Swin-transformer, Pvtv1, PVTv2, CrossViT, Dilateformer, EdgeNeXt, Efficientformer, FastViT, Mobile-former, ViT-Adapter, XCiT, ShuntedSA, PoolFormer, MPViT} have gradually become essential for computer vision tasks due to their ability to establish global perception by building long-range dependencies through self-attention.

Inspired by this, large kernels are increasingly recognized for their ability to establish long-range dependencies in CNNs, thereby improving the accuracy of various vision tasks.
The use of large kernels in convolutional networks originated with AlexNet \cite{AlexNet} and Inception \cite{Inception1,Inception2,Inceptionv4}, which utilized $7\times7$ or $11\times11$ kernels in the low-level layers.
ConvNeXt \cite{ConvNeXt} examines the feasibility of large-kernel convolution within conventional ResNet-like architectures, where performance saturates at a kernel size of $7\times7$.

Recently, RepLKNet \cite{RepLKNet} first scales the convolution kernel up to $31\times31$ using structural re-parameterization techniques and provides several guidelines for architectural design.
VAN \cite{VAN} acquires a $21\times{21}$ receptive field to establish long-term dependencies through sequential stacks of large-kernel depth-wise convolutions (DWConv) and depth-wise dilation convolutions.
SLaK \cite{SLaK} constructs a pure CNN architecture using sparse factorized $51\times{51}$ kernels, consisting of parallel $51\times{5}$ and $5\times{51}$ kernels.
MogaNet \cite{moganet} employs a spatial aggregation block utilizing $5\times{5}$ and $7\times{7}$ convolutions to adaptively aggregate discriminative features.
PeLK \cite{pelk} employs a human-like peripheral convolution to reduce parameters via parameter sharing while scaling the kernel size to a substantial $101\times{101}$.
However, recent large-kernel ConvNets suffer from parameters and FLOPs burden and disrupted AGD of ERF.
In this work, we focus on expanding ERF while maintaining AGD for ConvNets of Any Scale.
\section{Conclusion}

We introduce a Receptive Field Aggregator (RFA) to expand effective receptive field (ERF) while maintaining the Asymptotically Gaussian Distribution (AGD) of ERF.
Accordingly, we design a Three-layer RFA for input images with a resolution of $224\times{224}$, which can be a plug-and-play module for ConvNets or replace the convolutional layers within them.
Based on these designs, we propose a \textbf{\bfseries{uni}}versal \textbf{\bfseries{conv}}olutional neural \textbf{\bfseries{net}}work (ConvNet), termed UniConvNet, and evaluate its performance across a wide range of vision recognition tasks. 
All variants of UniConvNet demonstrate superior performance with reduced parameters and FLOPs.
This work may draw attention to the design of large ERF following an AGD, enhancing the ConvNet of any scale.

{
	\small
	\bibliographystyle{ieeenat_fullname}
	\bibliography{main}
}
\clearpage
\setcounter{page}{1}
\maketitlesupplementary


\appendix

\section*{Appendix}

\section{A proper Asymptotically Gaussian Distribution of small-scale pixels is more important than expanding the Effective Receptive Field}\label{Effective Receptive Field and Multi-scale perception ability}

\begin{figure}[t]
	\centering
	\includegraphics[width=0.9\linewidth]{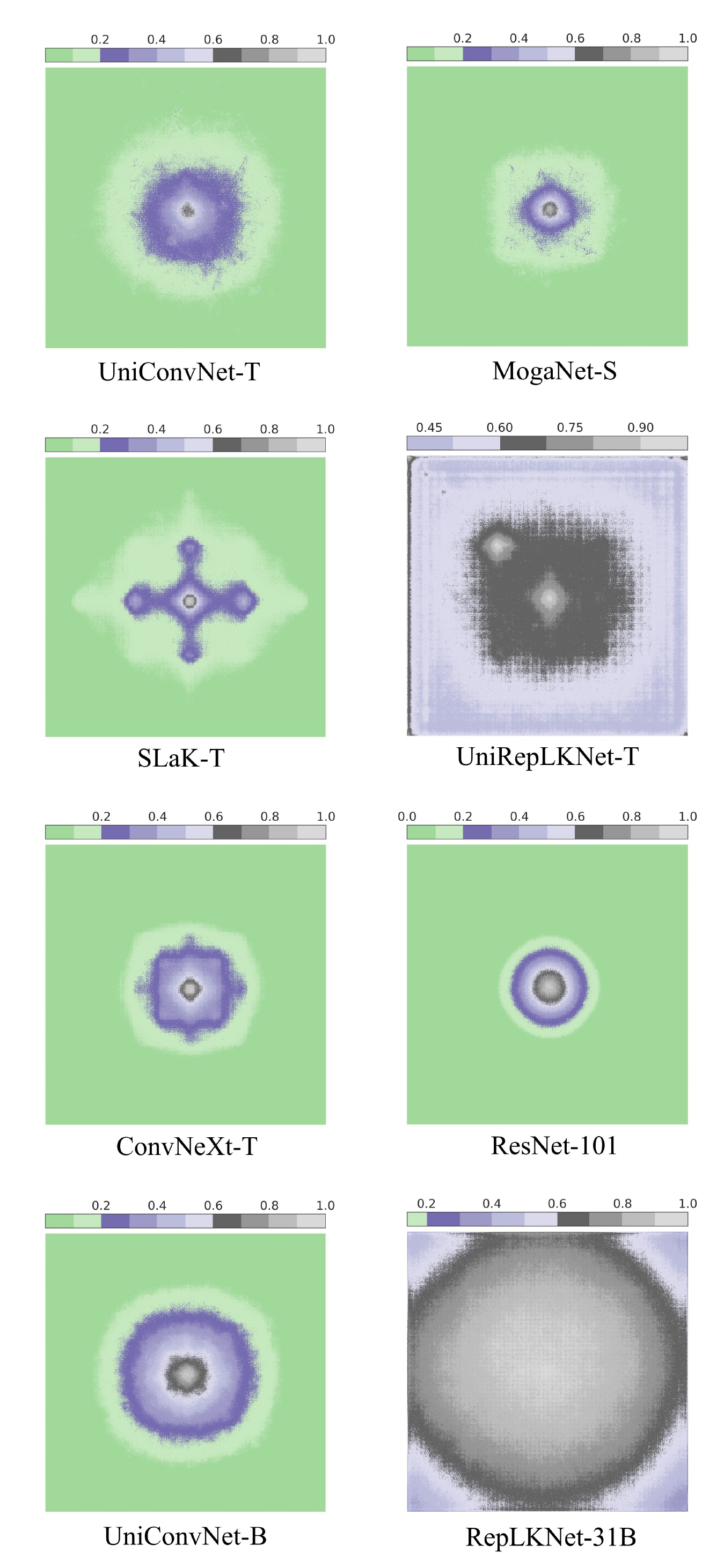}
	
	\caption{\textbf{\bfseries{Effective Receptive Field (ERF) of UniConvNet-T, MogaNet-S, SLaK-T, ConvNeXt-T,  UniRepLKNet-T, ResNet-101 and UniConvNet-B, RepLKNet-31B.}} The more stepped colour area indicates better AGD. The wider area indicates a larger ERF. Each ERF is based on an average of 1000 images with a resolution of $224\times{224}$.}
	\label{fig:Effective Receptive Field T}
\end{figure}

In dense prediction tasks (e.g., detection and segmentation), integrating contextual information via a large effective receptive field (ERF)  \cite{ERF} and distinguishing pixels of different scales are crucial.
As shown in \cref{fig:Effective Receptive Field T}, the ERF is used to visualize the effectiveness of the proposed Three-layer RFA.

MogaNet-S \cite{moganet} has similar asymptotically Gaussian distribution (AGD) at small-scale pixels, around the center, compared to UniConvNet-T, but UniConvNet-T exhibits a significantly larger ERF. 
This indicates that \textbf{\bfseries{expanding the ERF}} while \textbf{\bfseries{maintaining the AGD}} could help to generate a multi-scale impact, following AGD from center to edge, of a larger ERF, which consequently enhances the performance.

When comparing MogaNet-S \cite{moganet} with ConvNeXt-T \cite{ConvNeXt}, MogaNet-S \cite{moganet} has a better AGD at small-scale pixels with comparable ERF scale. 
This enables MogaNet-S \cite{moganet} to have superior performance, demonstrating that the \textbf{\bfseries{AGD of small-scale pixels}} is \textbf{\bfseries{more important}} when the \textbf{\bfseries{ERF scale is comparable}}.

Compared to UniConvNet-T, SLaK-T \cite{SLaK} achieves comparable ERF scale while disrupting the AGD of ERF.
The top-1 accuracy on ImageNet increased by $1.7\%$ point owing to the proper AGD on the area larger than small-scale area in SLaK-T \cite{SLaK}.
UniRepLKNet-T \cite{UniRepLKNet} achieves much larger ERF, compared to SLaK-T \cite{SLaK}, with inferior AGD, which benefits from extremely large ERF.
It is constrained by high parameters and FLOPs costs compared with UniConvNet-T.
This demonstrates that the sparsity \cite{SLaK} and re-parameterization \cite{DBB, repvgg} techniques effectively enlarge the ERF but suffer from improper AGD of smaller-scale pixels (the dark gray area in UniRepLKNet-T \cite{UniRepLKNet}).
Compared to UniConvNet-B, RepLKNet-31B \cite{RepLKNet} achieves a larger ERF but compromises small-scale AGD, with a $1.0\%$ TOP-1 accuracy drop on ImageNet. 
These phenomena typically demonstrate our viewpoint that \textbf{\bfseries{a proper asymptotically Gaussian distribution of small-scale pixels is more important than expanding the Effective Receptive Field}}.

\begin{figure*}[t]
	\centering
	\includegraphics[width=\linewidth]{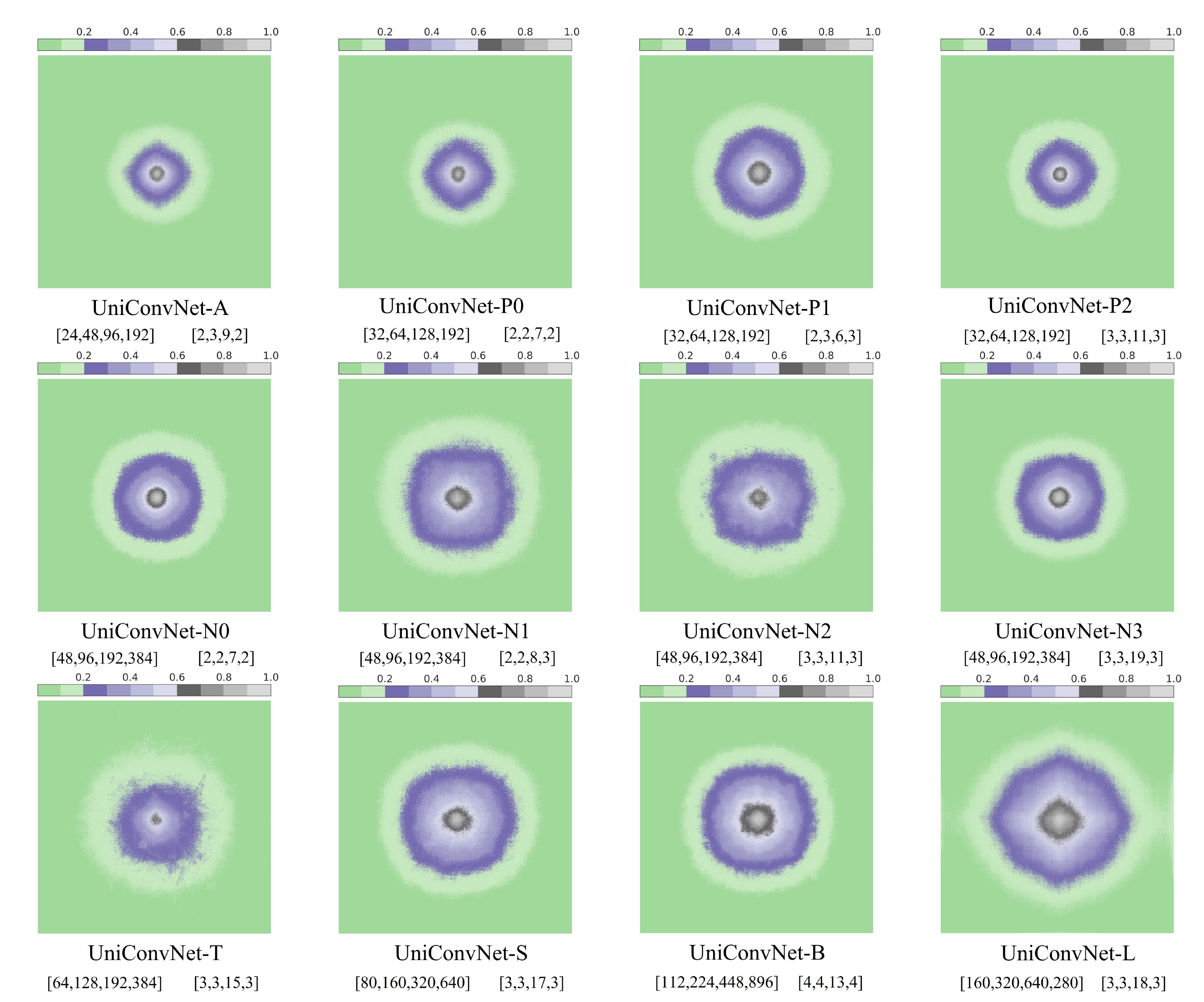}
	
	\caption{\textbf{\bfseries{Effective Receptive Field (ERF) of all UniConvNet variants.}} The more stepped colour area indicates better AGD. The wider area indicates a larger ERF. Each ERF is based on an average of 1000 images with a resolution of $224\times{224}$.}
	\label{fig:Effective Receptive Field ALL}
\end{figure*}

As shown in \cref{fig:Effective Receptive Field ALL}, UniConvNet variants consistently demonstrate large ERF while maintaining AGD.
These findings suggest that the Three-layer RFA can extend the ERF with proper combination of smaller kernels (e.g., $7\times{7}$, $9\times{9}$, $11\times{11}$). 

\section{Throughput Analysis Configuration}\label{Throughput analysis}
We use an A100 40GB GPU to benchmark throughput on several classic and relevant models. 
The software environment is PyTorch 1.13, CUDA 11.7, cuDNN 8.5. 
The hardware and software configurations align with InternImage \cite{DCNV3} to ensure a fair comparison.
The overall throughput of each model, measured as the number of images processed per second, is reported in FP32/FP16 data formats.

\section{Illustration of UniConvNet Block}
\label{Illustration of UniConvNet Block}

\subsection{Stem \& Downsampling Block}

Similar to ConvNeXt \cite{ConvNeXt} and InternImage \cite{DCNV3}, our model adopts a pyramid architecture with a stem block and three downsampling blocks to generate multi-scale feature maps. 
As shown in \cref{fig:Architecture}, The stem block, positioned before the first stage, reduces the input resolution by a factor of $4$. 
The stem block employs a bottleneck design, comprising two stacked $3\times3$ convolution layers and LayerNorm layers, interspersed with a GELU activation function to introduce nonlinearity to input images. 
The $3\times3$ convolutions have strides of $2$ and padding of $1$. The first convolution output channels are half those of the second. 
The downsampling block between stages reduces the input resolution by a factor of $2$. 
It consists of a LayerNorm layer followed by a $3\times3$ convolution with a stride of $2$ padding of $1$.

\subsection{Basic Block}

Inspired by the state-of-the-art CNN model InternImage, which integrates LayerNorm \cite{LayerNorm}, feed-forward networks \cite{Transformer}, and GELU \cite{GELU}, UniConvNet incorporates three stacked residual components in basic blocks.
Each residual component begins with a LayerNorm layer to normalize input features, followed sequentially by the Three-layer RFA, modified DCNV3 \cite{DCNV3}, and a feed-forward network.
The basic block initially employs the Three-layer RFA residual component to extract multi-scale features from small- and large-scale patterns, establishing long-range and multi-scale dependencies.
Following the Three-layer RFA residual component, a $3\times{3}$ (modified DCNV3 \cite{DCNV3}) residual convolution block and a feed-forward network are used for densely local perception, akin to conventional ConvNets. 

\section{Training Settings}

\begin{table}
	\belowrulesep=0pt
	\aboverulesep=0pt
	\centering
	\resizebox{\linewidth}{!}
	{%
		\begin{tabular}{@{}l|c|c|c@{}}
			\multirow{3}{8em}{\textbf{\bfseries{Super Prameters}}}
			& \textbf{\bfseries{UniConvNet-A}}
			& \textbf{\bfseries{UniConvNet-N0/N1}}
			& \multirow{2}{8em}{\textbf{\bfseries{UniConvNet-L/XL(S)}}}
			\\
			& \textbf{\bfseries{/P0/P1/P2(W)}}
			& \textbf{\bfseries{/N2/N3/T/S/B(S)}}
			&
			\\
			& ImageNet-1K
			& ImageNet-1K
			& ImageNet-22K
			\\
			\midrule
			Input Scale
			& 224\textsuperscript{2}
			& 224\textsuperscript{2}
			& 192\textsuperscript{2}
			\\
			Training Epochs
			& 300
			& 300
			& 90
			\\
			Batch Size
			& 4096
			& 4096
			& 4096
			\\
			\midrule
			Optimizer
			& AdamW
			& AdamW
			& AdamW
			\\
			Optimizer Momentum
			& $\beta_{1},\beta_{1}=0.9,0.999$
			& $\beta_{1},\beta_{1}=0.9,0.999$
			& $\beta_{1},\beta_{1}=0.9,0.999$
			\\
			Base Learning Rate
			& $4{e^{-3}}$
			& $4{e^{-3}}$
			& $4{e^{-3}}$
			\\
			Learning Rate Schedule
			& cosine
			& cosine
			& cosine
			\\
			Learning Rate Decay
			& $5{e^{-2}}$
			& $5{e^{-2}}$
			& $5{e^{-2}}$
			\\
			Layer-wise Learning Rate Decay
			& \usym{2613}
			& \usym{2613}
			& \usym{2613}
			\\
			Warmup Epochs
			& 20
			& 20
			& 20
			\\
			Warmup Schedule
			& linear
			& linear
			& linear
			\\
			\midrule
			Label Smoothing $\varepsilon$
			& 0.1
			& 0.1
			& 0.1
			\\
			Dropout Rate
			& \usym{2613}
			& \usym{2613}
			& \usym{2613}
			\\
			Drop Path Rate
			& 0.05/0.05/0.05/0.08
			& 0.08/0.1/0.1/0.1/0.2/0.4/0.6
			& 0.2/0.2
			\\
			Layer Scale
			& $1{e^{-6}}$
			& $1{e^{-6}}$
			& $1{e^{-6}}$
			\\
			\midrule
			RandAugment
			& (9,0.5)
			& (9,0.5)
			& (9,0.5)
			\\
			Color Jitter
			& 0.4
			& 0.4
			& 0.4
			\\
			Horizontal Flip
			& \usym{2613}
			& \usym{2613}
			& \usym{2613}
			\\
			Random Resized Crop
			& \usym{2613}
			& \usym{2613}
			& \usym{2613}
			\\
			Repeated Augment
			& \usym{2613}
			& \usym{2613}
			& \usym{2613}
			\\
			Head Init Scale
			& \usym{2613}
			& \usym{2613}
			& \usym{2613}
			\\
			Mixup Alpha
			& \usym{2613}
			& 0.8
			& 0.8
			\\
			Cutmix Alpha
			& \usym{2613}
			& 1.0
			& 1.0
			\\
			Erasing Probability
			& \usym{2613}
			& 0.25
			& 0.25
			\\
			\midrule
			Gradient Clip
			& \usym{2613}
			& \usym{2613}
			& \usym{2613}
			\\
			Loss
			& Cross Entropy
			& Cross Entropy
			& Cross Entropy
			\\
			Exp. Mov. Avg. (EMA)
			& 0.9999
			& 0.9999
			& \usym{2613}
			\\
		\end{tabular}
	}
	\caption{\textbf{\bfseries{(Pre-)Training settings for various model variants on ImageNet-1K/22K.}} The training recipes adhere to standard practices \cite{DeiT, ConvNeXt, DCNV3, DCNV4, FastViT, UniRepLKNet, EMO}, with certain tune-ups removed. 
	Multiple stochastic depth drop rates (e.g., 0.08/0.1/0.1/0.1/0.2/0.4/0.6) are assigned to UniConvNet-N0/N1/N2/N3/T/S/B, respectively.
	“W” and “S” indicate that the UniConvNet variants are trained using the weak and strong training recipes, respectively.}
	\label{tab:pretrain-settings}
\end{table}

\subsection{ImageNet-1K/22K Training}
\label{ImageNet-1K/22K Training Settings}

We adopt the commonly used training recipes from state-of-the-art methods \cite{DeiT, ConvNeXt, DCNV3, DCNV4, FastViT, UniRepLKNet, EMO} and remove some tune-ups for fair comparisons and to better represent the effectiveness of the proposed UniConvNet. 
Additionally, we apply a weak training recipe, following EMO \cite{EMO}, to improve performance on smaller models (UniConvNet-A/P0/P1/P2), and a strong training recipe, based on common practice \cite{ConvNeXt}, for larger variants (UniConvNet-N0/N1/N2/N3/T/S/B).
All experiments are conducted on the ImageNet-1K \cite{ImageNet} dataset, comprising $1000$ object classes and $1.2$ million training images.

Using the weak training recipe, we train UniConvNet-A/P0/P1/P2 models from scratch with $224\times{224}$ inputs for $300$ epochs.
The AdamW optimizer is used with a learning rate of $4 \times 10^{-3}$.
Training begins with a $20$-epoch linear warmup, followed by a cosine decay learning rate schedule.
A batch size of $4096$ and a weight decay of $0.05$ are employed.
RandAugment \cite{RandAugment} is applied for data augmentation in the weak training recipe.
Regularization techniques, including Stochastic Depth \cite{StochasticDepth} and Label Smoothing \cite{LabelSmoothing}, are employed.
A Layer Scale \cite{ LayerScale} with an initial value of $1 \times 10^{-6}$ is used.
Exponential Moving Average (EMA) \cite{EMA} is employed to reduce overfitting in larger models.

For UniConvNet-N0/N1/N2/N3/T/S/B, the strong training recipe is applied, incorporating additional data augmentation techniques such as Mixup \cite{Mixup}, Cutmix \cite{Cutmix}, and Random Erasing \cite{RandomErasing}, to enhancethe dataset for training on larger models.
For UniConvNet-L, we follow the strong training recipe and change the input image size to $192\times{192}$. 
Detailed training configurations for different model variants are provided in \cref{tab:pretrain-settings}.

\begin{table}
	\belowrulesep=0pt
	\aboverulesep=0pt
	\centering
	\resizebox{\linewidth}{!}
	{%
		\begin{tabular}{@{}l|c|c@{}}
			\multirow{2}{8em}{\textbf{\bfseries{Super Prameters}}}
			& \textbf{\bfseries{UniConvNet-T/S/B}}
			& \textbf{\bfseries{UniConvNet-L/XL}}
			\\
			& ImageNet-1K pt
			& ImageNet-22K pt
			\\
			
			& ImageNet-1K ft
			& ImageNet-1K ft
			\\
			\midrule
			Input Scale
			& 384\textsuperscript{2}
			& 384\textsuperscript{2}
			\\
			Training Epochs
			& 30
			& 30
			\\
			Batch Size
			& 512
			& 512
			\\
			\midrule
			Optimizer
			& AdamW
			& AdamW
			\\
			Optimizer Momentum
			& $\beta_{1},\beta_{1}=0.9,0.999$
			& $\beta_{1},\beta_{1}=0.9,0.999$
			\\
			Base Learning Rate
			& $5{e^{-5}}$
			& $5{e^{-5}}$
			\\
			Learning Rate Schedule
			& cosine
			& cosine
			\\
			Learning Rate Decay
			& $1{e^{-8}}$
			& $1{e^{-8}}$
			\\
			Layer-wise Learning Rate Decay
			& 0.7
			& 0.8
			\\
			Warmup Epochs
			& \usym{2613}
			& \usym{2613}
			\\
			Warmup Schedule
			& \usym{2613}
			& \usym{2613}
			\\
			\midrule
			Label Smoothing $\varepsilon$
			& 0.1
			& 0.1
			\\
			Dropout Rate
			& \usym{2613}
			& \usym{2613}
			\\
			Drop Path Rate
			& 0.4/0.6/0.8
			& 0.3/0.35
			\\
			Layer Scale
			& pre-trained
			& pre-trained
			\\
			\midrule
			RandAugment
			& (9,0.5)
			& (9,0.5)
			\\
			Color Jitter
			& 0.4
			& 0.4
			\\
			Horizontal Flip
			& \usym{2613}
			& \usym{2613}
			\\
			Random Resized Crop
			& \usym{2613}
			& \usym{2613}
			\\
			Repeated Augment
			& \usym{2613}
			& \usym{2613}
			\\
			Head Init Scale
			& 0.001
			& 0.001
			\\
			Mixup Alpha
			& \usym{2613}
			& \usym{2613}
			\\
			Cutmix Alpha
			& \usym{2613}
			& \usym{2613}
			\\
			Erasing Probability
			& 0.25
			& 0.25
			\\
			\midrule
			Gradient Clip
			& \usym{2613}
			& \usym{2613}
			\\
			Loss
			& Cross Entropy
			& Cross Entropy
			\\
			Exp. Mov. Avg. (EMA)
			& 0.9999
			& 0.9999
			\\
		\end{tabular}
	}
	\caption{\textbf{\bfseries{Fine-tuning settings for various model variants on ImageNet-1K.}} The training recipe follows common practices \cite{ConvNeXt, UniRepLKNet}. Multiple stochastic depth drop rates (e.g., 0.4/0.6/0.8) are for UniConvNet-T/S/B, respectively. “ImageNet-1K pt”, “ImageNet-1K ft”, and “ImageNet-22K pt” represent ImageNet-1K pre-training, ImageNet-1K fine-tuning and ImageNet-22K pre-training, respectively.}
	\label{tab:fine-tune-settings}
\end{table}

\subsection{ImageNet-1K Fine-tuning}
\label{ImageNet-1K Fine-tune Settings}

For ImageNet-1K fine-tuning, compared to the strong training recipes for ImageNet-1K/22K, the base learning rate of the AdamW optimizer is set to $5 \times 10^{-5}$.
The learning rate decay is set to $1 \times 10^{-8}$.
ImageNet-1K fine-tuning is performed with a batch size of $512$, without requiring warm-up.
Different layer-wise learning rate decay factors are applied: $0.7$ for UniConvNet-T/S/B and $0.8$ for UniConvNet-L.
Data augmentation techniques, Mixup \cite{Mixup} and Cutmix \cite{Cutmix}, are removed to improve fine-tuning results.
Additionally, ImageNet-1K pre-trained UniConvNet-T/S/B and ImageNet-22K pre-trained UniConvNet-L are fine-tuned at an increased resolution of $384 \times {384}$.

\begin{table}
	\belowrulesep=0pt
	\aboverulesep=0pt
	\centering
	\resizebox{0.75\linewidth}{!}
	{%
		\begin{tabular}{@{}l|c|c|c@{}}
			UniConvNet & 
			\#Params & 
			ACC-W(\%) & 
			ACC-S(\%)
			\\
			\midrule
			
			UniConvNet-A & 
			3.4M &
			\textbf{\bfseries{77.0}} &
			\usym{2613}
			\\
			\midrule
			
			UniConvNet-P0 & 
			5.2M &
			\textbf{\bfseries{79.1}} &
			\usym{2613}
			\\
			\midrule
			
			UniConvNet-P1 &  
			6.1M &
			\textbf{\bfseries{79.6}} &
			78.8
			\\
			\midrule
			
			UniConvNet-P2 &  
			7.6M &
			\textbf{\bfseries{80.5}} &
			79.9
			\\
			\midrule
			
			\rowcolor{blue!12}
			\textbf{\bfseries{UniConvNet-N0}} &  
			10.2M &
			\textbf{\bfseries{81.6}} &
			\textbf{\bfseries{81.6}}
			\\
			\midrule
			
			\rowcolor{blue!12}
			\textbf{\bfseries{UniConvNet-N1}} &  
			13.1M &
			81.8 &
			\textbf{\bfseries{82.2}}
			\\
			\midrule
			
			UniConvNet-N2 &
			15.0M &
			\usym{2613} &
			\textbf{\bfseries{82.7}}
			\\
		\end{tabular}
	}
	\caption{\textbf{\bfseries{Explorations of Training Recipes for Classification.}} “ACC-W(\%)” and “ACC-S(\%)” are the TOP-1 accuracy trained by weak traininig recipe and strong traininig recipe, respectively.}
	\label{tab:Table-Selection-Training-recipe}
\end{table}

\subsection{Training Recipes for Classification}
\label{Selection-Training-recipe}

We evaluate the performance of UniConvNet-A/P0/P1/P2/\\N0/N1/N2 on ImageNet-1K to conduct an ablation study comparing weak and strong training recipes.
As illustrated in \cref{tab:Table-Selection-Training-recipe}, the weak training recipe exhibits overfitting with models having $10.2M$ parameters, while models with $13.2M$ parameters start benefiting from the strong training recipe.
Consequently, the choice of training recipes is straightforward: UniConvNet-A/P0/P1/P2/N0 adopts the weak training recipe to leverage smaller-scale datasets.
UniConvNet-N1/N2 and larger models employ the strong training recipe, utilizing augmented \cite{Mixup, Cutmix, RandomErasing} datasets to optimize performance with larger parameter sizes.

\subsection{Object Detection and Instance Segmentation Fine-tuning}
\label{Object Detection and Instance Segmentation Fine-tune}

Following EMO \cite{EMO}, we utilize the standard MMDetection \cite{MMDetection} library and the AdamW \cite{AdamW} optimizer to train the heavy RetinaNet and light SSDLite models with a batch size of $16$ on $8$ A100 GPUs.

Following standard practices \cite{ConvNeXt,DCNV3,DCNV4}, we further fine-tune the scaled-up UniConvNet using batch sizes of $16$ and $8$, respectively, for fair comparisons.
Under the $1\times$ schedule, images are resized so that the shorter side is $800$ pixels and the longer side does not exceed $1333$ pixels.
During testing, the shorter side is fixed at $800$ pixels.
Under the $3\times$ schedule, the longer side remains capped at $1333$ pixels, while the shorter side is resized to a range of $480$–$800$ pixels.
We also employ the standard MMDetection \cite{MMDetection} library and the AdamW \cite{AdamW} optimizer for training, using a base learning rate of $1 \times 10^{-4}$.

\subsection{Semantic Segmentation Fine-tuning}
\label{Semantic Segmentation Fine-tune}

We fine-tune DeepLabv3 \cite{DeepLabv3} and PSPNet \cite{PSPNet} using the ImageNet-1K pre-trained UniConvNet on the ADE20K \cite{ADE20K} dataset.
Following EMO \cite{EMO}, we use the MMSegmentation \cite{MMSegmentation} library and the AdamW \cite{AdamW} optimizer to train DeepLabv3 \cite{DeepLabv3} and PSPNet \cite{PSPNet} for 160k iterations on 8 A100 GPUs, ensuring fair comparisons.

The scaled-up UniConvNet is fine-tuned with the UperNet framework on ADE20K for 160k iterations, using a batch size of $16$.
The AdamW \cite{AdamW} optimizer is used for training.
The base learning rates are set to $6\times{10^{-5}}$ for UniConvNet-N2/N3/T/S/B and $2\times{10^{-5}}$ for UniConvNet-L.
A polynomial decay schedule with a power of $1.0$ is applied for learning rate decay.
Following common practices \cite{Swin-transformer,ConvNeXt,DCNV3, DCNV4}, images are cropped to $512 \times 512$ for UniConvNet-N2/N3/T/S/B and $640 \times 640$ for UniConvNet-L to ensure fair comparisons.

\end{document}